\documentclass[10pt,twocolumn,letterpaper]{article}
\usepackage[table]{xcolor}
\usepackage{indentfirst}
\usepackage{amsfonts}
\usepackage{pifont}
\usepackage{circledsteps}
\usepackage{makecell}
\usepackage{multirow}
\usepackage{booktabs}
\usepackage{caption}
\usepackage{amsmath}
\usepackage{cancel}
\usepackage{pifont}
\usepackage{bbding}
\usepackage{bm}
\usepackage{circledsteps}
\usepackage[accsupp]{axessibility}

\usepackage[review]{cvpr}      

\definecolor{cvprblue}{rgb}{0.21,0.49,0.74}
\definecolor{hrefpink}{HTML}{ED06A1} 
\usepackage[pagebackref,breaklinks,colorlinks,allcolors=cvprblue]{hyperref}

\title{\textbf{An End-to-End Robust Point Cloud Semantic Segmentation Network with Single-Step Conditional Diffusion Models}}

\makeatletter
\renewcommand\thanks[1]{\footnotemark[\arabic{footnote}]\protected@xdef\@thanks{\@thanks
        \protect\footnotetext[\arabic{footnote}]{#1}}}
\makeatother

\author{Wentao Qu$^1$,
Jing Wang$^2$, 
YongShun Gong$^3$,
Xiaoshui Huang$^{4*}$,
Liang Xiao$^{1*}$\\
NJUST$^1$, THU‌$^2$, SDU$^3$, SJTU$^4$\\
{\tt\small\small $\{$quwentao, jwang$\}$@njust.edu.cn,
huangxiaoshui@163.com, xiaoliang@mail.njust.edu.cn}
\thanks{$^*$Corresponding Author. \href{https://github.com/QWTforGithub/CDSegNet}{\textcolor{hrefpink}{https://github.com/QWTforGithub/CDSegNet}}}
}

\nolinenumbers

\date{}

\begin{document}
\maketitle




\begin{abstract}
Existing conditional Denoising Diffusion Probabilistic Models (DDPMs) with a Noise-Conditional Framework (NCF) remain challenging for 3D scene understanding tasks, as the complex geometric details in scenes increase the difficulty of fitting the gradients of the data distribution (the scores) from semantic labels. This also results in longer training and inference time for DDPMs compared to non-DDPMs. From a different perspective, we delve deeply into the model paradigm dominated by the Conditional Network. In this paper, we propose an end-to-end robust semantic \textbf{Seg}mentation \textbf{Net}work based on a \textbf{C}onditional-Noise Framework (CNF) of D\textbf{D}PMs, named \textbf{CDSegNet}. Specifically, CDSegNet models the Noise Network (NN) as a learnable noise-feature generator. This enables the Conditional Network (CN) to understand 3D scene semantics under multi-level feature perturbations,  enhancing the generalization in unseen scenes. Meanwhile, benefiting from the noise system of DDPMs, CDSegNet exhibits  strong robustness for data noise and sparsity in experiments. Moreover, thanks to CNF, CDSegNet can generate the semantic labels in a single-step inference like non-DDPMs, due to avoiding directly fitting the scores from semantic labels in the dominant network of CDSegNet. On public indoor and outdoor benchmarks, CDSegNet significantly outperforms existing methods, achieving state-of-the-art performance.
\end{abstract}


\begin{figure}[htp]
	\centering
	\includegraphics[width=0.48\textwidth]{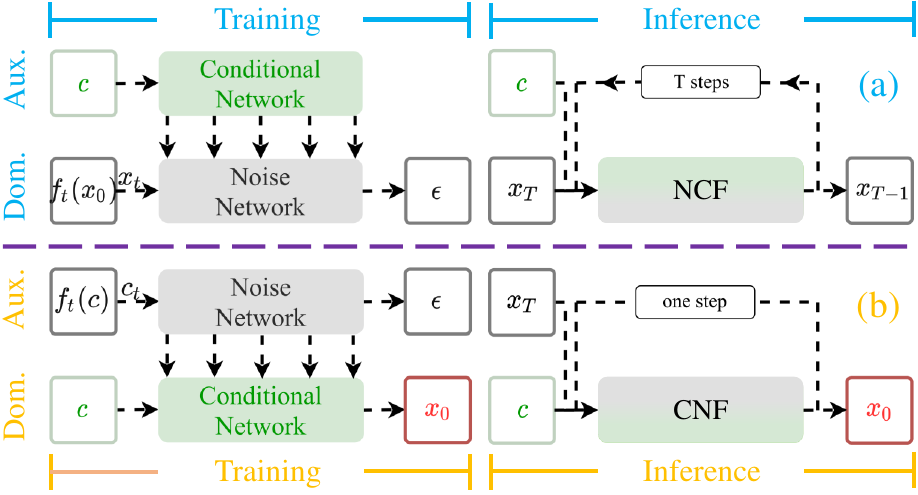}
 \vspace{-0.8cm}
	\caption{The training and inference difference between NCF and CNF. NCF dominated by NN, relies on the noise fitting quality, requiring extensive training and inference iterations. In contrast, CNF alleviates the noise fitting necessity by focusing on CN, cleverly avoiding this issue,  alongside retaining the DDPM robustness.}
	\label{fig_cnf_ncf}

\end{figure}

\section{Introduction}
Point cloud, as a fundamental 3D representation, provides the most crucial data structure support for 3D tasks. Benefiting from the rapid development of 3D devices and continuous innovation in data synthesis techniques, large-scale scene point clouds have become accessible \cite{dai2017scannet, armeni20163d, caesar2020nuscenes, fong2022panoptic}. Therefore, accurate semantic understanding of 3D scenes, applied to a wide range of 3D downstream tasks such as autonomous driving \cite{huang2022imfnet}, robotic technology \cite{huang2022gmf}, and virtual reality \cite{kharroubi2019classification}, has gained increasing attention.


Encouraged by deep learning, a large number of learnable point cloud semantic segmentation methods have achieved significant results in recent years \cite{zhao2021point, choy20194d, lai2022stratified, wang2023octformer, wu2022point, wu2024point}. Nevertheless, these methods often overlook the fact that raw point clouds from 3D devices are usually perturbed and sparse \cite{rozenberszki2022language, caesar2020nuscenes}, making them sensitive to data noise and sparsity \cite{xiang2019generating, huang2022frozen, shi2022shape}. This limits the optimal segmentation accuracy, especially recognizing in object boundaries and small objects \cite{ye2021learning, ye2022robust}.


Along another research line, DDPMs \cite{ho2020denoising} with a strong noise-robust denoising architecture, originated from and flourished in image generation \cite{dhariwal2021diffusion, rombach2022high, lin2023magic3d}, have been explored in various 3D tasks \cite{zhou20213d, lyu2021conditional, qu2024conditional}. They typically consider the 3D task as a conditional generation problem, built upon the Noise-Conditional Framework (NCF, see Fig.~\ref{fig_cnf_ncf}(a)). In general, the Conditional Network (CN) extracts the conditional features for generating guidance. Meanwhile, the Noise Network (NN) predicts the scores \cite{song2020score} from task targets, dominating the results of tasks. 


Unfortunately, this framework poses challenges when applied to real-time 3D scene understanding tasks, such as autonomous driving. This is because, to better fit the scores from task targets, DDPMs typically require more training and inference steps compared to non-DDPMs \cite{song2020denoising, song2020score}. Although some methods can accelerate the  DDPM sampling process  \cite{song2020denoising, lu2022dpm}, they still require dozens of steps, alongside a sub-optimization process. Moreover, the complex distribution of 3D scenes makes it difficult for DDPMs to achieve satisfactory results in an end-to-end manner. (see Fig.~\ref{fig_training_inference})

To address the above problems, in this paper, we \textbf{rethink} the existing end-to-end framework of conditional DDPMs, initially revealing some key insights into the DDPM advantages and limitations in point cloud semantic segmentation. 

Inspired by these core insights, we design a  Conditional-Noise Framework (CNF, see Fig.~\ref{fig_cnf_ncf}(b)) of DDPMs, effectively preserving advantages and circumventing limitations. Unlike NCF, CNF treats NN and CN as the auxiliary network and the dominant network in 3D tasks, respectively. In this way, NN in CNF can still inherit the noise construction system of DDPMs, thus preserving the noise and sparsity robustness from DDPMs. Meanwhile, this also enables NN in CNF to relax the requirement of fitting the scores from task targets. Therefore, CNF allows the model to follow the DDPM pattern in training but not in inference, overcoming the demand for extensive iterations from DDPMs.




Furthermore, we propose an end-to-end robust point cloud semantic segmentation network based on CNF, called CDSegNet. Specifically, CDSegNet treats NN as a lightweight noise-feature generator. The noise information from NN, effectively filtered via a Feature Fusion Module (FFM), reasonably perturbs the semantic features in CN. This motivates the generalization ability of CDSegNet in unknown scenes \cite{wang2023better, luo2020learn, rebuffi2021data}. Meanwhile, under CNF, CDSegNet can follow the noise addition pattern of DDPMs during training, maintaining the robustness to noise from the modeled distribution \cite{qu2024conditional} and the robustness to sparse scenes and insufficient data. Moreover, benefiting from the dominance of CN in CNF, CDSegNet can be regarded as a non-DDPM during inference. This enables CDSegNet to produce the semantic labels in a single-step inference. 






To the best of our knowledge, we are the first end-to-end attempt to introduce DDPMs into point cloud semantic segmentation \cite{liu20243d, zheng2024point}. We lower the threshold and encourage more researchers to further explore applications of DDPMs to 3D tasks. Our key contributions can be summarized as:
\begin{itemize}
    \item We systematically analyze and identify the advantages and limitations of DDPMs with a Noise-Conditional Framework in point cloud semantic segmentation, offering new knowledge for DDPMs in 3D tasks.
    \item We design a Conditional-Noise Framework of DDPMs, preserving strengths while circumventing shortcomings.
    \item We propose an end-to-end robust point cloud segmentation network based on CNF, CDSegNet, exhibiting strong robustness and requiring only a single-step inference.
    \item Comprehensive experiments on large-scale indoor and outdoor benchmarks demonstrate that CDSegNet achieves significant performance and strong robustness.
   
\end{itemize}

\section{Related Works}
\textbf{Learnable Point Cloud Semantic Segmentation.} Benefiting from the powerful data-driven capability of deep learning, directly extracting features from point clouds to understand 3D scene semantics has become possible \cite{qi2017pointnet, qi2017pointnet++, choy20194d, thomas2019kpconv}. Inspired by the aforementioned, a multitude of methods have emerged in recent years, achieving significant success in point cloud semantic segmentation. Early methods usually utilize RNNs to establish interactions between point cloud slices, attempting to improve feature extraction effectiveness through building ordered relationships within point clouds \cite{huang2018recurrent, ye20183d}. However, the substantial computational cost greatly limits the input scale. To resolve this problem, some researchers focus on large-scale scene segmentation \cite{landrieu2018large, hu2020randla, fan2021scf}. Although remarkable progress has been made, the methods are still constrained by a small receptive field, limiting the further improvement in segmentation results. Lately, some methods built upon Transformers have been proposed. These methods inherit the capability of modeling long-range dependencies, overcoming the limitations of the feature receptive field and thereby achieving state-of-the-art results \cite{guo2021pct, zhao2021point, wu2022point, lai2022stratified, yang2023swin3d, wu2024point}.

Although existing methods focusing on segmentation accuracy have achieved impressive results, they overlook the fact that raw point clouds often exhibits perturbed and sparse. This leads to them sensitive to data noise and sparsity. In this paper, we introduce DDPMs with a Conditional-Noise Framework to address the above problem. This demonstrates the strong robustness to data noise and sparsity, while avoiding extensive training and inference steps.

\textbf{DDPMs for 3D Tasks.} DDPMs, succeeded in image generation, have conducted some explorations in various 3D tasks. This typically transforms the 3D task as a conditional generation problem. \cite{luo2021diffusion} first introduces DDPMs into point cloud generation, providing inspiration for subsequent explorations. Then, some works attempt to extend DDPMs to point cloud completion \cite{zhou20213d, lyu2021conditional}. Subsequently, \cite{qu2024conditional} undertakes preliminary investigations into DDPMs for point cloud upsampling. Furthermore, several works have integrated DDPMs into point cloud semantic segmentation using a pre-training (two-stage training) approach \cite{liu20243d, zheng2024point}.

Although some explorations demonstrate the potential of DDPMs in 3D tasks, the two-stage training requirement in scene semantic understanding tasks demonstrate that DDPMs still face the challenge of fitting the scores in complex 3D scenes. Meanwhile, the dozens or even thousands of inference steps limit practical applications in 3D tasks with real-time requirements. In this paper, we propose an end-to-end robust point cloud semantic segmentation network based on our CNF of DDPMs. Thanks to CNF, our method circumvents directly fitting the scores from semantic labels in the dominant segmentation network, requiring only a single-step inference.

\section{Conditional DDPMs for PCSS}


In this section, we first introduce conditional DDPMs. Next, we identify the reason behind leveraging conditional DDPMs for point cloud semantic segmentation (PCSS) and discuss the limiting factors associating with this approach.

\subsection{Background}
\label{3.1}




Given a target data $\bm{x_0} \sim P_{data}$, a guiding condition $\bm{c} \sim P_{c}$ and a latent variable $\bm{z} \sim P_{noise}$, conditional DDPMs follow an auto-regressive process \cite{song2020denoising, qu2024conditional}: a predefined diffusion process $q$ that gradually destroys the data content until $\bm{x_0}$ degrades into $\bm{z}$, and a trainable conditional generation process $p_\theta$ that slowly generates the specific result until $\bm{z}$ is recovered to $\bm{x_0}$ under the guidance of the condition $\bm{c}$. We can apply the framework to multiple 3D tasks \cite{luo2021diffusion, lyu2021conditional, qu2024conditional}. In PCSS, $\bm{x_0}$ represents the target semantic label, while $\bm{c}$ means the segmented point cloud.

We consider the noise as the fitting target, due to the better performance observed in experiments \cite{ho2020denoising}. Then, the training objective under specific conditions is \cite{qu2024conditional}:

\vspace{-15pt}
\begin{equation}
\begin{split}
	\label{f311}
	L(\theta) =
 \mathbb{E}_{\bm{\epsilon} \sim \mathcal{N}(0,I)}||\bm{\epsilon} - \bm{\epsilon_\theta}(\bm{x_t},C)||^2, 
\end{split}
\end{equation}
where $C$=$\{\bm{c},t\}$ represents the conditions, while $t \sim \mathcal{U}(T)$ ($T$=1000). Therefore, unconditional generation ($\bm{c}=\emptyset$) can be viewed as a special case of conditional DDPMs conditioned on the time label $t$ for controlling the noise level. That is, our analysis is generalizable to any type of DDPMs.






Meanwhile, under stochastic differential equations (SDEs), the target noise in conditional DDPMs can be converted into and from the score (the gradient of the data distribution) by a constant factor $\alpha = -\frac{1}{\sqrt{1-\overline{\alpha}_t}}$ \cite{song2021scorebased} :

\vspace{-13pt}
\begin{equation}
\begin{split}
	\label{f312}
	\alpha \bm{\epsilon_\theta}(\bm{x_t},C) = s_\theta\bm{(x_t},C) \approx \nabla_{\bm{x_t}} \log P_t(\bm{x_t}).
\end{split}
\end{equation}


\subsection{What Supports the Use of DDPMs in PCSS?}
\label{3.2}

Point clouds in real-world scenes are often noisy and sparse \cite{rozenberszki2022language, caesar2020nuscenes}. Existing methods  overlook this fact, making them sensitive to data  noise  and sparsity \cite{xiang2019generating, huang2022frozen, shi2022shape}. In this paper, we introduce DDPMs to address this issue.

\textbf{Noise robustness.} Benefiting from the noise system, \textbf{\textit{DDPMs inherently present the robustness to noise from the modeled distribution}}  \cite{qu2024conditional}. We further reveal the key sources of the robustness in the system: \textbf{\textit{the noise samples and the noise fitting}}. According to Eq.~\ref{f311},  DDPMs can access multi-level noise samples $\bm{x_t}$ from the modeled distribution and use the standard noise $\bm{\epsilon}$ as the fitting target. This facilitates the model to understand the task information under modeled distribution perturbations, enhancing the adaptability to the relevant distribution noise. Sec.~\ref{5.3} further discusses and supports the conclusion.

\textbf{Sparsity robustness.}  \textbf{\textit{DDPMs  exhibit better performance on under-sampled data (sparse scenes and less data)}}. In fact, this noise-adding manner can be formally regarded as a kind of \textbf{\textit{data augmentation}} \cite{xiang2023denoising,wang2023better,chen2024deconstructing}: 

\vspace{-10pt}
\begin{equation}
\begin{split}
	\label{f321}
	L(\theta) =
 \mathbb{E}_{\bm{\epsilon} \sim \mathcal{N}(0,I)}||\bm{\epsilon} - \bm{\epsilon_\theta}(f_t(\bm{x_0}),\bm{C})||^2, 
\end{split}
\end{equation}
where $f_t(\bm{x_0}) = \sqrt{1-\overline{\alpha}_t} \bm{\epsilon} + \sqrt{\overline{\alpha}_t}\bm{x_0}$ \cite{ho2020denoising}. The linear function $f_t(\cdot)$ maps $\bm{x_0}$ to a more ambiguous latent distribution.

This data augmentation can significantly alleviate the model overfitting (see Tab.~\ref{tab561} and Fig.~\ref{fig_loss_range}), enhancing the generalization for outdoor sparse scenes (Sec.~\ref{5.1}, Sec.~\ref{5.2} and Sec.~\ref{5.5}) and less data (Sec.~\ref{5.4}).



\subsection{What Limits the Use of  DDPMs in PCSS?}
\label{3.3}

\begin{figure}[htp]
	\centering
	\includegraphics[width=0.48\textwidth]
 {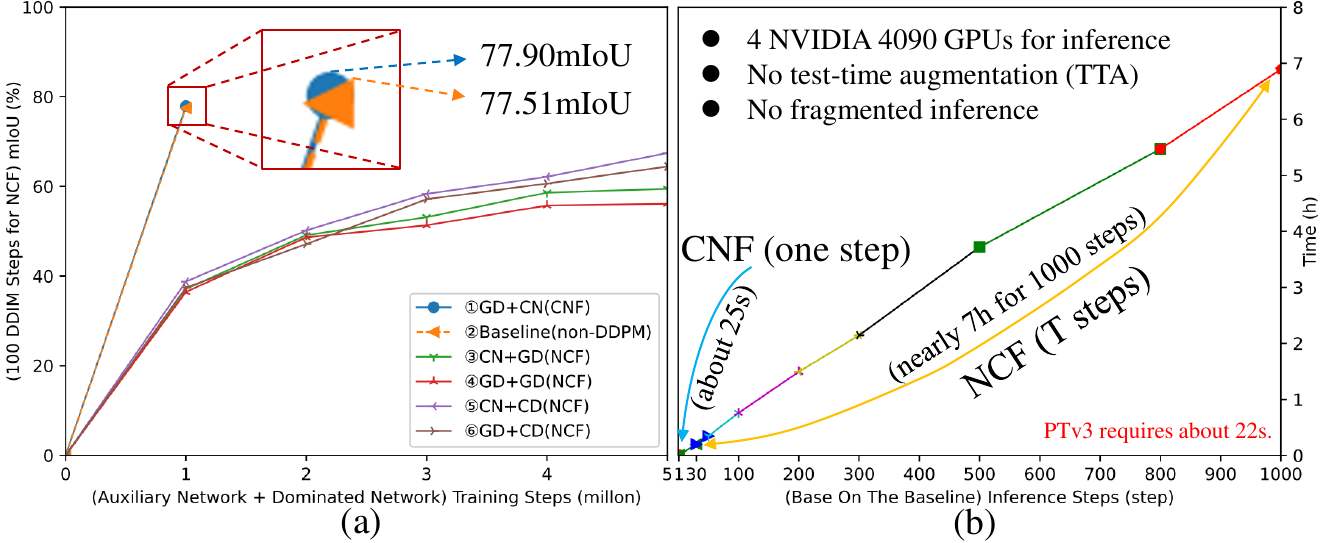}
 \vspace{-0.8cm} 
	\caption{We try several combinations for conditional DDPMs built on the baseline (Sec.~\ref{5.1}) on ScanNet in (a). GD+CD (NCF) indicates that NN and CN are modeled as  Gaussian \cite{ho2020denoising} and  categorical \cite{austin2021structured} diffusion  (details in the supplementary material). To better fit noise, these combinations dominated by NN (\small \textcircled{3},\small \textcircled{4},\small \textcircled{5},\small \textcircled{6}) require more iterations to converge than CNF, but exhibiting poorer performance, due to complex scene distribution. (b) shows the inference time cost of CNF and NCF under the  same baseline. CNF achieves better performance with fewer iterations.}
	\label{fig_training_inference}
\end{figure}

Unfortunately, DDPMs may hardly be applied to tasks with high real-time requirements, due to the extensive training and inference iterations they demand (see Fig.~\ref{fig_training_inference}).




As described in Eq.~\ref{f311}, the performance of DDPMs essentially lies in \textbf{\textit{the noise fitting quality}} (the proof in the supplementary material). To better approximate the noise target, \textbf{\textit{DDPMs require more training and inference iterations than non-DDPMs}}, due to the significant error of fitting distributions with a large difference in one step \cite{song2020score, song2020denoising, lu2022dpm}.

We provide a proof, under a unified setting for DDPMs and non-DDPMs, DDPMs require more steps $T$ to converge. For a semantic segmentation task, given a network $f_\theta$ with sufficient fitting ability and a semantic sample pair $(\bm{c}, \bm{x_0})$, the training objective of DDPMs and non-DDPMs can be consistently formulated as (assuming using MSE):

\vspace{-15pt}
\begin{equation}
\begin{split}
	\label{f331}
	L_{\theta} =\frac{1}{T}\sum_{t=1}^{T}||\bm{y_{t-1}}-f_\theta(\bm{x_{t}}, \bm{c})||^2,
\end{split}
\end{equation}
where the target $\bm{y_{t-1}}$ means the target noise conditioned on the time label $t$ in DDPMs.  Meanwhile, the input $\bm{x_t}$=$\bm{\mu_t}+\bm{\sigma_t}\bm{\epsilon}$ \cite{ho2020denoising},  and we omit the time label $t$ as part of the input.


The inference process commonly follows:

\vspace{-15pt}
\begin{equation}
\begin{split}
	\label{f332}
	 \bm{y_{t-1}^{'}}=f_\theta(\bm{x_{t}},\bm{c}), \quad t \in [1,T], 
\end{split}
\end{equation}
where $\bm{y_{t-1}^{'}}$ means the predicted noise in DDPMs.

According to Eq.~\ref{f331} and Eq.~\ref{f332}, the sufficient convergence for DDPMs requires at least the step size $T$$>$1 to accommodate all $\bm{y_{t-1}}$ (training) and $\bm{y_{t-1}^{'}}$ (inference), due to the significant error in one step. However, non-DDPMs only necessitate the step size $T$=1 (the input $\bm{x_{t}}$=$\emptyset$):

\vspace{-15pt}
\begin{equation}
\begin{split}
	\label{f333}
	L_{\theta} =||\bm{y_{0}}-f_\theta(\emptyset, \bm{c})||^2, \quad \bm{y_{0}^{'}}=f_\theta(\emptyset,\bm{c}),
\end{split}
\end{equation}
where the target $\bm{y_{0}}$=$\bm{x_0}$, while $\bm{y_{0}^{'}}$ means the predicted $\bm{x_0}$.


\section{Methodology}


\subsection{Conditional-Noise Framework} 
\label{4.1}
To maintain the advantage sources  (Sec.~\ref{3.2}) and circumvent the limitation factors (Sec.~\ref{3.3}), we design a Conditional-Noise Framework (CNF) of DDPMs (see Fig.~\ref{fig_cnf_ncf}(b)). Unlike the Noise-Conditional Framework (NCF), CNF regards the Conditional Network (CN, $f_\psi$)  as the dominant task backbone and uses the Noise Network (NN, $\epsilon_\theta$) as a auxiliary feature augmentation branch.

Specifically, to perturb the features in CN, the condition $\bm{c}$ is added with noise instead of the target $\bm{x_0}$ in NN:

\vspace{-13pt}
\begin{equation}
\begin{split}
	\label{f411}
	L(\theta) =
 \mathbb{E}_{\bm{\epsilon} \sim \mathcal{N}(0,I)}||\bm{\epsilon} - \epsilon_\theta(\bm{c_t},t)||^2. 
\end{split}
\end{equation}

Next, a Feature Fusion Module (FFM) is used to filter the noise information from NN, ensuring the feature perturbations in a reasonable manner.

Furthermore, CN learns the task-related information under multi-level feature perturbations via FFM: 

\vspace{-13pt}
\begin{equation}
\begin{split}
	\label{f412}
	L(\psi) =l_{task}(\bm{x_{0}}, f_\psi( f_t(\bm{c}), \bm{c})),
\end{split}
\end{equation}
where $f_t(\bm{c})$=$\mathcal{FFM}(\bm{c_{t}^{noise}})$ means a nonlinear function. $\mathcal{FFM}(\cdot)$ represents the Feature Fusion Module. $\bm{c_{t}^{noise}}$ means the noise feature from $\epsilon_\theta$ with $\bm{c_t}$ as input. $l_{task}(\cdot)$ indicates the task-related loss function. 


This simple and effective model paradigm:
\begin{itemize} 
    \item \textbf{Following Noise Construction System.} Eq.~\ref{f411} indicates that the noise system of DDPMs is still retained in NN, aligning with Eq.~\ref{f311}, maintaining the noise robustness.
    \item \textbf{Aligning with Feature Augmentation.} Eq.~\ref{f412} shows that CN follows the feature augmentation pattern, aligning with Eq.~\ref{f321}, thereby enhancing the sparsity robustness for sparse scenes and less data.
    \item \textbf{Relaxing Noise Fitting Requirement.} Eq.~\ref{f411} and Eq.~\ref{f412} mean that the model performance relies on CN rather than NN, relaxing the noise fitting requirement, avoiding excessive training and inference iterations from DDPMs. 
\end{itemize}



Moreover, NN in CNF allows us to transcend the limitation of modeling Gaussian diffusion \cite{ho2020denoising}. This means that CNF can use any distribution of DDPMs \cite{bansal2024cold} (see Fig.~\ref{fig_training_inference}).

\begin{figure}[htp]
	\centering
	\includegraphics[width=0.48\textwidth]
 {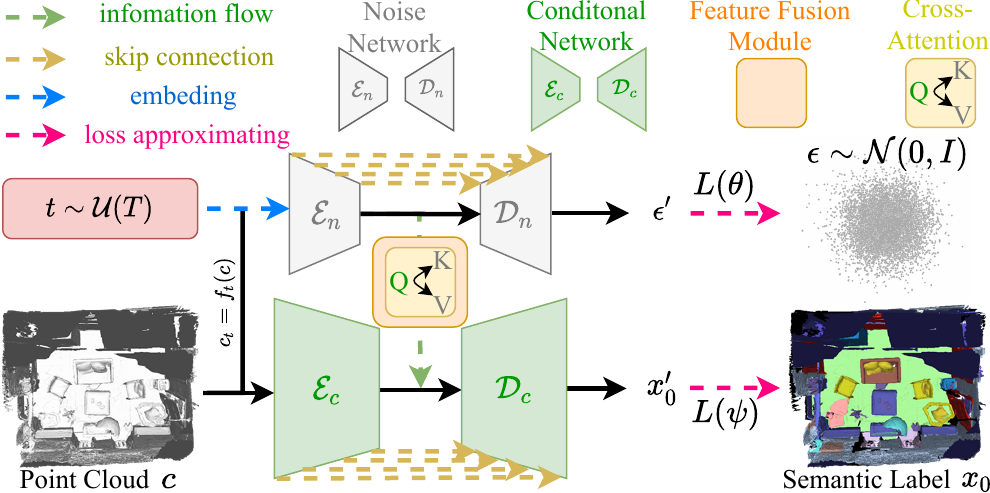}
 \vspace{-0.4cm} 
	\caption{The overall framework of CDSegNet. The auxiliary Noise Network (NN),  seen as a noise-feature generator, modeling the diffusion process conditioned on the time label, perturbs the input features at different noise levels. Meanwhile, the Feature Fusion Module (FFM) controls the noise information flow direction, achieving the semantic feature augmentation by reasonably filtering the perturbations. Furthermore, the dominant Conditional Network (CN) predicts the segmentation results in a pure manner. }
	\label{fig_overall}
\end{figure}

\vspace{-5pt}
\subsection{Network Architecture}
\label{4.2}

In this section, we introduce the overall architecture of CDSegNet aligned with CNF. This consists of three primary components: the auxiliary Noise Network (NN), the Feature Fusion Module (FFM), and the dominant Conditional Network (CN), as clearly illustrated in Fig.~\ref{fig_overall} (the parameter and optimization details in the supplementary material). 

\textbf{The Auxiliary Noise Network.} NN  constructs the noise system of DDPMs, perturbing the conditional point cloud. As mentioned in Sec.~\ref{4.1}, NN should exhibit lighter, due to the insignificant noise fitting requirement. This follows the Transformer-U-Net architecture \cite{zhao2021point, wu2022point}, stacking two-stage standard Transformer blocks in the encoder and decoder. Meanwhile, similar to \cite{wu2022point, wu2024point}, we utilize the effective grid pooling to achieve upsampling and downsampling. Moreover, the insignificant noise fitting also means that introducing the time label $t$ in NN is sufficient to model the diffusion process without considering additional conditions.



\textbf{The Feature Fusion Module.} FFM directs the information flow from NN to CN at the bottleneck stage. In fact, FFM can adaptively filter the noise information, making the feature augmentation from NN in a reasonable way, as excessive  perturbations may harm the  performance of CN.


Specifically, FFM first performs the feature projection via MLPs, $F_{cn} \in \mathbb{R}^{N_{cn} \times C_{cn}} \rightarrow (Q) \in \mathbb{R}^{N_{cn} \times C}$, $F_{nn} \in \mathbb{R}^{N_{nn} \times C_{nn}} \rightarrow (K,V) \in \mathbb{R}^{N_{nn} \times C}$. Subsequently, FFM filters the noise information via a Cross-Attention block \cite{vaswani2017attention}:

\begin{figure*}[htp]
	\centering
	\includegraphics[width=\textwidth]
 {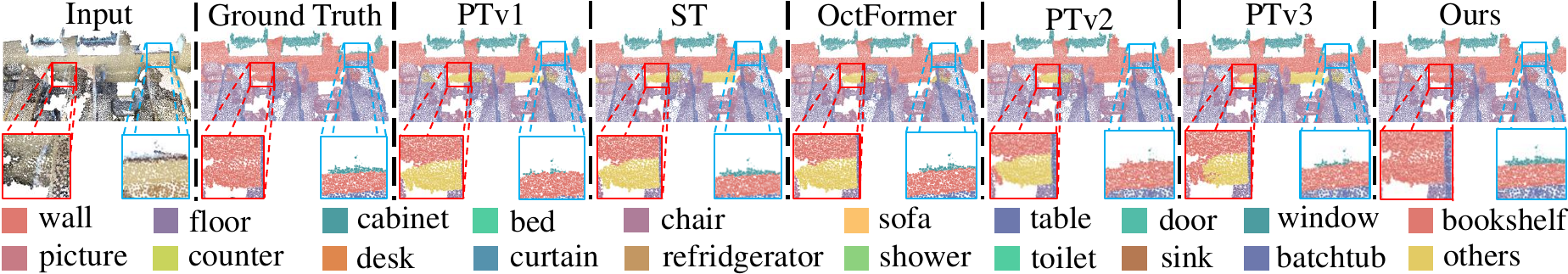}
 \vspace{-0.6cm} 
	\caption{The visualization results on ScanNet. Our method achieves better semantic segmentation results in object boundaries, such as the boundary between the wall and the wall (red solid box) and the boundary between the wall and the window (blue solid box). }
	\label{fig_scannet}
\end{figure*}

\begin{figure*}[htp]
	\centering
	\includegraphics[width=\textwidth]
 {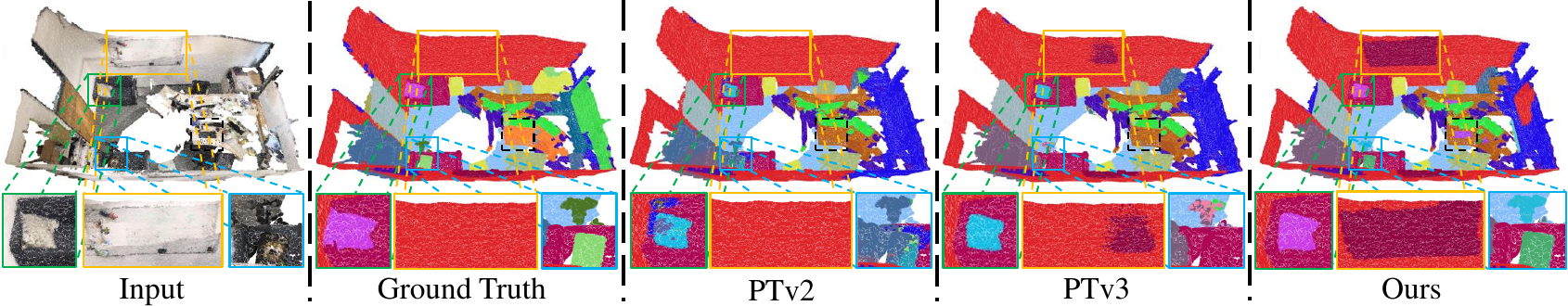}
 \vspace{-0.6cm} 
	\caption{The visualization results on ScanNet200. Our method achieves the clearer segmentation results in small object recognition, and segments the categories not labeled in the Ground Truth, such as the keyboard (black dashed box) and the whiteboard (orange solid box).}
	\label{fig_scannet200}
\end{figure*}

\vspace{-10pt}
\begin{equation}
\begin{split}
	\label{f421}
        O = mlp(WV)+F_{cn}, \\
        F = ffn(O)+O, \;\;
\end{split}
\end{equation}
where $W \in \mathbb{R}^{N_{cn} \times N_{nn}} = softmax(\frac{QK^T}{\sqrt{C}})$. 


We consider the unidirectional flow $F_{nn} \stackrel{FFM}{\longrightarrow} F_{cn}$, due to the better trade-off between the performance and computational cost and the sufficient diffusion modeling in NN.

\textbf{The Dominant Conditional Network.} CN follows the architecture of NN, with four stages for both the encoder and decoder, focusing on the segmentation results. Meanwhile, besides the feature perturbations from NN, we use only the conditional point cloud as input, ensuring that CN concentrates purely on the scene semantic understanding.

\subsection{Training and Inference}
\label{4.3}
\textbf{Training.} Following \cite{wu2022point, wu2024point}, we constrain the semantic segmentation results in CN:

\vspace{-10pt}
\begin{equation}
\begin{split}
	\label{f431}
        L(\psi) = CE(\bm{x_0}, \bm{x_0^{'}}) + \lambda Lovaz(\bm{x_0}, \bm{x_0^{'}}),
\end{split}
\end{equation}
where $CE(\cdot)$ represents the cross-entropy loss, while $Lovasz(\cdot)$ follows \cite{berman2018lovasz}. $\lambda=1$ means a weighting factor.

Furthermore, we can optimize CNF from a multi-task perspective (\textbf{the noise fitting in NN} and \textbf{the semantic label fitting in CN}). This applies the Geometric Loss Strategy (GLS) \cite{chennupati2019multinet++}, with a geometric mean weight, to alleviate the convergence speed difference between NN and CN, regulating the perturbations from NN in a reasonable manner:

\vspace{-10pt}
\begin{equation}
\begin{split}
	\label{f432}
        L_{total} = \prod_{i=1}^{n} \sqrt[n]{L_i} =\sqrt{L(\theta)} \cdot  \sqrt{ L(\psi)},
\end{split}
\end{equation}
where $n=2$ means the number of tasks.

\textbf{Inference.} Thanks to CNF, CDSegNet can be seen a non-DDPM during inference, requesting only one step:

\begin{equation}
    \label{f433}
    \bm{x_0^{'}}=\mathcal{D}_c(\mathcal{E}_c(\bm{c}),\mathcal{FFM}(\mathcal{E}_n(\bm{c_T},T))),
\end{equation}
where $\bm{c_T} \sim \mathcal{N}(\bm{0},\bm{I})$, while $\bm{x_0^{'}}$  means the predicted semantic label. 


\section{Experiments}

\subsection{Experiment Setup}
\label{5.1}

\textbf{Dataset.} Two indoor benchmarks (ScanNet \cite{dai2017scannet}, ScanNet200 \cite{rozenberszki2022language}) and one outdoor benchmark (nuScenes \cite{caesar2020nuscenes}) are used for evaluations. For ScanNet and ScanNet200, we divide train/val/test with 1201/312/100 scenes like \cite{wu2022point, wu2024point}. Meanwhile, the official protocol is followed to split train/val/test into 700/150/150 scenes for nuScenes. The point clouds in ScanNet, ScanNet200 and nuScenes are voxelized into 0.02m, 0.02m and 0.05m, respectively.

\textbf{Baseline.} To demonstrate the effectiveness of our method, we set a baseline. This eliminates the diffusion modeling in NN of CDSegNet, converting NN into a lightweight CN that approximates the input using MSE.

\begin{figure*}[htp]
	\centering
	\includegraphics[width=\textwidth]
 {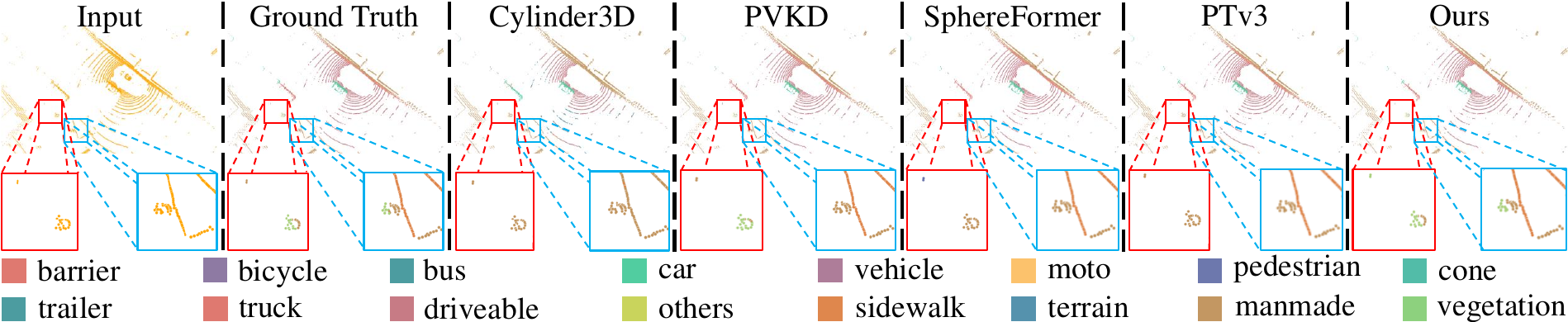}
 \vspace{-0.6cm} 
	\caption{The visualization results on nuScenes. Our method produces remarkable results in small object recognition, such as the vegetation (red solid box and blue solid box). }
	\label{fig_nuscenes}
\end{figure*}

\subsection{Comparison of Segmentation Results} 
\label{5.2}

\textbf{Indoor Dataset.} We first conduct evaluations on indoor datasets. Tab.~\ref{tab521} shows that CDSegNet achieves significant performance on ScanNet and ScanNet200. Meanwhile, CDSegNet outperforms all other methods across all metrics on ScanNet. This is because, benefiting from the strong noise robustness, CDSegNet can achieve better results on object boundaries with more perturbations. Fig.~\ref{fig_scannet} shows the visualization on ScanNet, further supporting the viewpoint. Moreover, CDSegNet also demonstrates significant results in recognizing small objects within more complex and perturbed scenes. As shown in Fig.~\ref{fig_scannet200}, CDSegNet can more clearly identify small objects, such as the pillow (green solid box) and the whiteboard (orange solid box).

Actually, as shown in Fig.~\ref{fig_scannet200}, the Ground Truth sometimes includes incorrect annotations (e.g., some points of sofa are labeled as the pillow category) or may even omit annotations (e.g., the whiteboard and the keyboard). This may be one of the reasons why most models perform poorly on ScanNet200. Meanwhile, the significant results on mIoU indicate that our CDSegNet can achieve better performance in perturbed environments compared to other methods.


\vspace{-2pt}
\begin{table}[h]
        \scriptsize
  \resizebox{0.48\textwidth}{!}{
	\begin{tabular}{p{1.6cm}p{0.6cm}p{0.6cm}p{0.7cm}p{0.005cm}p{0.6cm}p{0.6cm}p{0.7cm}}	
        \Xhline{1pt}
  
        \multirow{2}{*}{Methods}
        &\multicolumn{3}{c}{ScanNet \cite{dai2017scannet}} 
        &\quad
        &\multicolumn{3}{c}{ScanNet200 \cite{rozenberszki2022language}} \\
         \cline{2-4} \cline{6-8}

        &\makecell[c]{mIoU}
        &\makecell[c]{mAcc}
        &\makecell[c]{allAcc}
        &\quad
        &\makecell[c]{mIoU}
        &\makecell[c]{mAcc}
        &\makecell[c]{allAcc}\\
       \hline

        PTv1 \cite{zhao2021point}
        &\makecell[c]{70.8}
        &\makecell[c]{76.4}
        &\makecell[c]{87.5}
        &\quad
        &\makecell[c]{29.8}
        &\makecell[c]{40.5}
        &\makecell[c]{78.4}\\

        MinkUNet \cite{choy20194d}
        &\makecell[c]{72.3}
        &\makecell[c]{79.4}
        &\makecell[c]{89.1}
        &\quad
        &\makecell[c]{28.3}
        &\makecell[c]{39.8}
        &\makecell[c]{77.9}\\

        ST \cite{lai2022stratified}
        &\makecell[c]{74.3}
        &\makecell[c]{82.5}
        &\makecell[c]{90.7}
        &\quad
        &\makecell[c]{-}
        &\makecell[c]{-}
        &\makecell[c]{-}\\

        OctFormer \cite{wang2023octformer}
        &\makecell[c]{75.0}
        &\makecell[c]{83.1}
        &\makecell[c]{91.3}
        &\quad
        &\makecell[c]{32.9}
        &\makecell[c]{42.4}
        &\makecell[c]{81.2}\\

        PTv2 \cite{wu2022point}
        &\makecell[c]{75.5}
        &\makecell[c]{82.9}
        &\makecell[c]{91.2}
        &\quad
        &\makecell[c]{31.4}
        &\makecell[c]{42.0}
        &\makecell[c]{80.7}\\

        PTv3 \cite{wu2024point}
        &\makecell[c]{\underline{77.6}}
        &\makecell[c]{85.0}
        &\makecell[c]{\underline{92.0}}
        &\quad
        &\makecell[c]{35.3}
        &\makecell[c]{\textbf{46.0}}
        &\makecell[c]{83.3}\\

        Baseline
        &\makecell[c]{77.5}
        &\makecell[c]{\underline{85.1}}
        &\makecell[c]{91.9}
        &\quad
        &\makecell[c]{\underline{35.4}}
        &\makecell[c]{45.5}
        &\makecell[c]{\underline{83.6}}\\

        \rowcolor{gray!20} 
        Ours
        &\makecell[c]{\textbf{77.9}}
        &\makecell[c]{\textbf{85.2}}
        &\makecell[c]{\textbf{92.2}}
        &\quad
        &\makecell[c]{\textbf{36.3}}
        &\makecell[c]{\underline{45.9}}
        &\makecell[c]{\textbf{83.9}}\\

        \Xhline{1pt}
        
	\end{tabular}
 }
	\caption{The results on ScanNet and ScanNet200. Our method surpasses other methods in almost all metrics.}
	\label{tab521}

\end{table}
\vspace{-8pt}

\textbf{Outdoor Dataset.} We also conduct the validation on the outdoor benchmark. Compared to indoor scenes, the spatial distance between points in outdoor point clouds is larger. This causes more significant data sparsity.  Tab.~\ref{tab522} shows that CDSegNet exhibits the excellent results, significantly outperforming existing methods on the mIoU. As mentioned in Sec.~\ref{4.1}, CNF aligns with a feature augmentation strategy,  improving   generalization on sparse scenes. Fig.~\ref{fig_nuscenes} further demonstrates the superiority of CDSegNet in small object recognition in large outdoor sparse scenes, such as the vegetation (red solid box and blue solid box).

\begin{table*}[h]
  \resizebox{1.0\textwidth}{!}{
        \begin{tabular}{p{2.9cm}|p{0.9cm}|p{1.0cm}p{1.1cm}p{0.9cm}p{0.9cm}p{1.2cm}p{0.9cm}p{1.5cm}p{0.9cm}p{1.0cm}p{1.0cm}p{1.2cm}p{1.0cm}p{1.3cm}p{1.0cm}p{1.4cm}p{1.5cm}}	
        \Xhline{1pt}
  
        {Methods}
        &\makecell[c]{mIoU}
        &{\makecell[c]{barrier}}
        &{\makecell[c]{bicycle}}
        &{\makecell[c]{bus}}
        &{\makecell[c]{car}}
        &{\makecell[c]{vehicle}}
        &{\makecell[c]{moto}}
        &{\makecell[c]{pedestrian}}
        &{\makecell[c]{cone}}
        &{\makecell[c]{trailer}}
        &{\makecell[c]{truck}}
        &{\makecell[c]{drivable}}
        &{\makecell[c]{others}}
        &{\makecell[c]{sidewalk}}
        &{\makecell[c]{terrain}}
        &{\makecell[c]{manmade}}
        &{\makecell[c]{vegetation}} \\
        
       \hline

        RangeNet53++ \cite{milioto2019rangenet++}
        &\makecell[c]{65.5}
        &\makecell[c]{66.0}
        &\makecell[c]{21.3}
        &\makecell[c]{77.2}
        &\makecell[c]{80.9}
        &\makecell[c]{30.2}
        &\makecell[c]{66.8}
        &\makecell[c]{69.6}
        &\makecell[c]{52.1}  
        &\makecell[c]{54.2}  
        &\makecell[c]{72.3}  
        &\makecell[c]{94.1}  
        &\makecell[c]{66.6} 
        &\makecell[c]{63.5} 
        &\makecell[c]{70.1} 
        &\makecell[c]{83.1}
        &\makecell[c]{79.8}
        \\
     
        PolarNet \cite{zhang2020polarnet}
        &\makecell[c]{71.0}
        &\makecell[c]{74.7}
        &\makecell[c]{28.2}
        &\makecell[c]{85.3}
        &\makecell[c]{90.9}
        &\makecell[c]{35.1}
        &\makecell[c]{77.5}
        &\makecell[c]{71.3}
        &\makecell[c]{58.8}  
        &\makecell[c]{57.4}  
        &\makecell[c]{76.1}  
        &\makecell[c]{96.5}  
        &\makecell[c]{71.1} 
        &\makecell[c]{74.7} 
        &\makecell[c]{74.0} 
        &\makecell[c]{87.3}
        &\makecell[c]{85.7}\\

        Salsanext \cite{cortinhal2020salsanext}
        &\makecell[c]{72.2}
        &\makecell[c]{74.8}
        &\makecell[c]{34.1}
        &\makecell[c]{85.9}
        &\makecell[c]{88.4}
        &\makecell[c]{42.2}
        &\makecell[c]{72.4}
        &\makecell[c]{72.2}
        &\makecell[c]{63.1}
        &\makecell[c]{61.3}
        &\makecell[c]{76.5}
        &\makecell[c]{96.0}
        &\makecell[c]{70.8}
        &\makecell[c]{71.2}
        &\makecell[c]{71.5}
        &\makecell[c]{86.7}
        &\makecell[c]{84.4}
        \\

        AMVNet \cite{liong2012amvnet}
        &\makecell[c]{76.1}
        &\makecell[c]{79.8}
        &\makecell[c]{32.4}
        &\makecell[c]{82.2}
        &\makecell[c]{86.4}
        &\makecell[c]{\textbf{62.5}}
        &\makecell[c]{81.9}
        &\makecell[c]{75.3}
        &\makecell[c]{\textbf{72.3}}
        &\makecell[c]{83.5}
        &\makecell[c]{65.1}
        &\makecell[c]{\textbf{97.4}}
        &\makecell[c]{67.0}
        &\makecell[c]{\textbf{78.8}}
        &\makecell[c]{74.6}
        &\makecell[c]{90.8}
        &\makecell[c]{87.9}\\

        Cylinder3D \cite{zhu2021cylindrical}
        &\makecell[c]{76.1}
        &\makecell[c]{76.4}
        &\makecell[c]{40.3}
        &\makecell[c]{91.2}
        &\makecell[c]{\underline{93.8}}
        &\makecell[c]{51.3}
        &\makecell[c]{78.0}
        &\makecell[c]{78.9}
        &\makecell[c]{64.9}
        &\makecell[c]{62.1}
        &\makecell[c]{84.4}
        &\makecell[c]{96.8}
        &\makecell[c]{71.6}
        &\makecell[c]{76.4}
        &\makecell[c]{75.4}
        &\makecell[c]{90.5}
        &\makecell[c]{87.4}
        \\

        PVKD \cite{hou2022point}
        &\makecell[c]{76.0}
        &\makecell[c]{76.2}
        &\makecell[c]{40.0}
        &\makecell[c]{90.2}
        &\makecell[c]{\textbf{94.0}}
        &\makecell[c]{50.9}
        &\makecell[c]{77.4}
        &\makecell[c]{78.8}
        &\makecell[c]{64.7}
        &\makecell[c]{62.0}
        &\makecell[c]{84.1}
        &\makecell[c]{96.6}
        &\makecell[c]{71.4}
        &\makecell[c]{76.4}
        &\makecell[c]{\underline{76.3}}
        &\makecell[c]{90.3}
        &\makecell[c]{86.9}
        \\

        RPVNet \cite{xu2021rpvnet}
        &\makecell[c]{77.6}
        &\makecell[c]{78.2}
        &\makecell[c]{43.4}
        &\makecell[c]{92.7}
        &\makecell[c]{93.2}
        &\makecell[c]{49.0}
        &\makecell[c]{85.7}
        &\makecell[c]{80.5}
        &\makecell[c]{66.0}
        &\makecell[c]{66.9}
        &\makecell[c]{84.0}
        &\makecell[c]{96.9}
        &\makecell[c]{73.5}
        &\makecell[c]{75.9}
        &\makecell[c]{76.0}
        &\makecell[c]{90.6}
        &\makecell[c]{88.9}
        \\

        SphereFormer \cite{lai2023spherical}
        &\makecell[c]{79.5}
        &\makecell[c]{78.7}
        &\makecell[c]{46.7}        
        &\makecell[c]{95.2}
        &\makecell[c]{{93.7}}
        &\makecell[c]{54.0}
        &\makecell[c]{88.9}
        &\makecell[c]{81.1}
        &\makecell[c]{68.0}
        &\makecell[c]{\textbf{74.2}}
        &\makecell[c]{\textbf{86.2}}
        &\makecell[c]{\underline{97.2}}
        &\makecell[c]{74.3}
        &\makecell[c]{{76.3}}
        &\makecell[c]{75.8}
        &\makecell[c]{\textbf{91.4}}
        &\makecell[c]{\textbf{89.7}}        
        \\


        PTv3 \cite{wu2024point}
        &\makecell[c]{80.3}
        &\makecell[c]{\textbf{80.5}}
        &\makecell[c]{\textbf{53.8}}     
        &\makecell[c]{\underline{95.9}}
        &\makecell[c]{91.9}
        &\makecell[c]{52.1}
        &\makecell[c]{88.9}
        &\makecell[c]{\textbf{84.5}}
        &\makecell[c]{\underline{71.7}}
        &\makecell[c]{\underline{74.1}}
        &\makecell[c]{{84.5}}
        &\makecell[c]{\underline{97.2}}
        &\makecell[c]{\underline{75.6}}
        &\makecell[c]{77.0}
        &\makecell[c]{{76.2}}
        &\makecell[c]{{91.2}}
        &\makecell[c]{\underline{89.6}}  \\

        Baseline
        &\makecell[c]{\underline{80.4}}
        &\makecell[c]{\underline{80.1}}
        &\makecell[c]{53.2}        
        &\makecell[c]{{\underline{95.9}}}
        &\makecell[c]{92.0}
        &\makecell[c]{56.5}
        &\makecell[c]{\underline{89.2}}
        &\makecell[c]{{84.1}}
        &\makecell[c]{{71.2}}
        &\makecell[c]{{73.1}}
        &\makecell[c]{{84.4}}
        &\makecell[c]{96.9}
        &\makecell[c]{\textbf{76.5}}
        &\makecell[c]{{77.2}}
        &\makecell[c]{{75.8}}
        &\makecell[c]{\underline{91.3}}
        &\makecell[c]{89.4}  \\

        \rowcolor{gray!20} 
        Ours
        &\makecell[c]{\textbf{81.2}}
        &\makecell[c]{\underline{80.1}}
        &\makecell[c]{\underline{53.5}}        
        &\makecell[c]{\textbf{97.0}}
        &\makecell[c]{92.3}
        &\makecell[c]{\underline{62.3}}
        &\makecell[c]{\textbf{89.7}}
        &\makecell[c]{\underline{84.2}}
        &\makecell[c]{\underline{71.7}}
        &\makecell[c]{72.2}
        &\makecell[c]{\underline{85.9}}
        &\makecell[c]{\underline{97.2}}
        &\makecell[c]{\textbf{76.5}}
        &\makecell[c]{\underline{77.8}}
        &\makecell[c]{\textbf{76.9}}
        &\makecell[c]{\textbf{91.4}}
        &\makecell[c]{\textbf{89.7}}  \\

        \Xhline{1pt}
        
	\end{tabular}
 }
	\caption{The results on nuScenes. Our method significantly outperforms other methods in
the term of mIoU.}
	\label{tab522}
\end{table*}

\subsection{Validation for Noise Robustness}
\label{5.3}

We further validate the noise robustness of CDSegNet. This adds a Gaussian perturbation $\bm{n_G} \sim \mathcal{N}(\bm{n_G};\bm{0},\tau\bm{I})$ to the normalized inputs of models
\cite{he2023grad, qu2024conditional}, i.e., $\bm{c'}=\bm{c} + \bm{n_G}$. 

\begin{figure}[htp]
	\centering
	\includegraphics[width=0.48\textwidth]{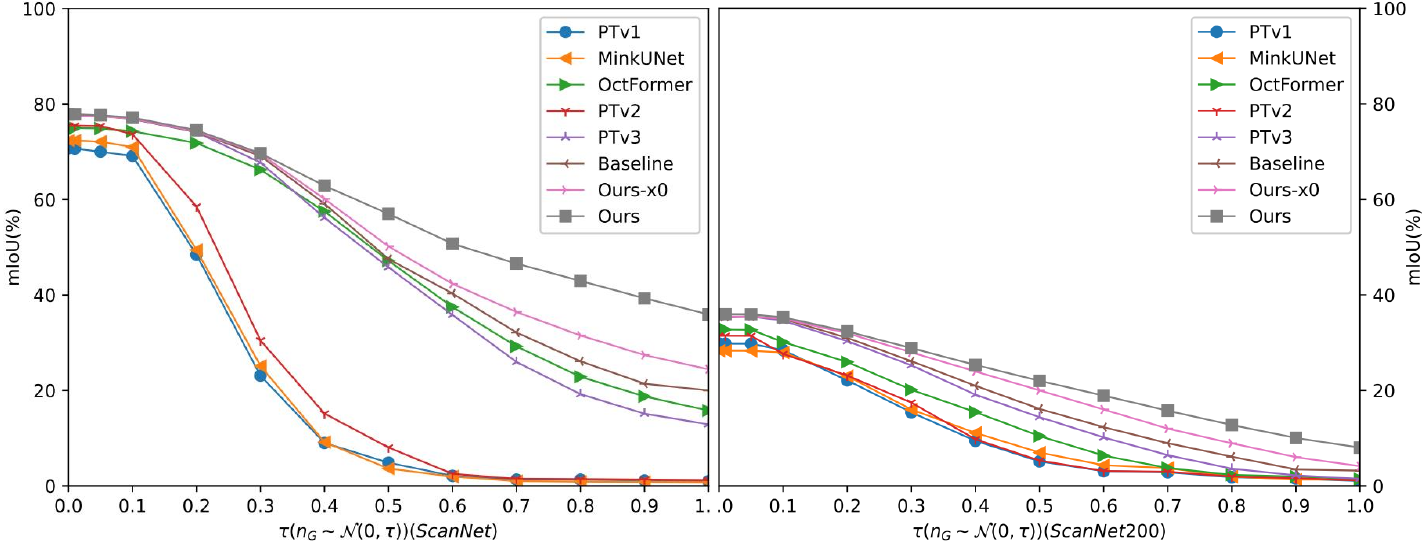}
 \vspace{-0.7cm} 
	\caption{The noise robustness results on ScanNet and ScanNet200. Ours-$\bm{x_0}$ means the fitting target is $\bm{x_0}$ in NN of CDSegNet. Our method consistently  outperforms all other methods.}
	\label{fig_noise_robustness}

\end{figure}

Fig.~\ref{fig_noise_robustness} demonstrates the strong noise robustness of CDSegNet on ScanNet and ScanNet200. Meanwhile, we can observed that Ours and Ours-$x_0$ are stronger than the baseline, validating the conclusion in Sec.~\ref{3.2}. However, Ours-$x_0$ performs significantly weaker than Ours. Although Ours-$x_0$ retains the noise sample $\bm{x_t}$ during training, the fitting target is $\bm{x_0}$, not $\bm{\epsilon}$. This leads to the model lacking the further noise adaptability in scenes. We believe that \textbf{the noise fitting in the noise system of DDPMs contributes mainly to the noise robustness} (we hope to validate the conclusion in the future).




\vspace{-5pt}
\vspace{-6pt}

\vspace{-10pt}
\begin{figure}[htp]
	\centering
	\includegraphics[width=0.48\textwidth]{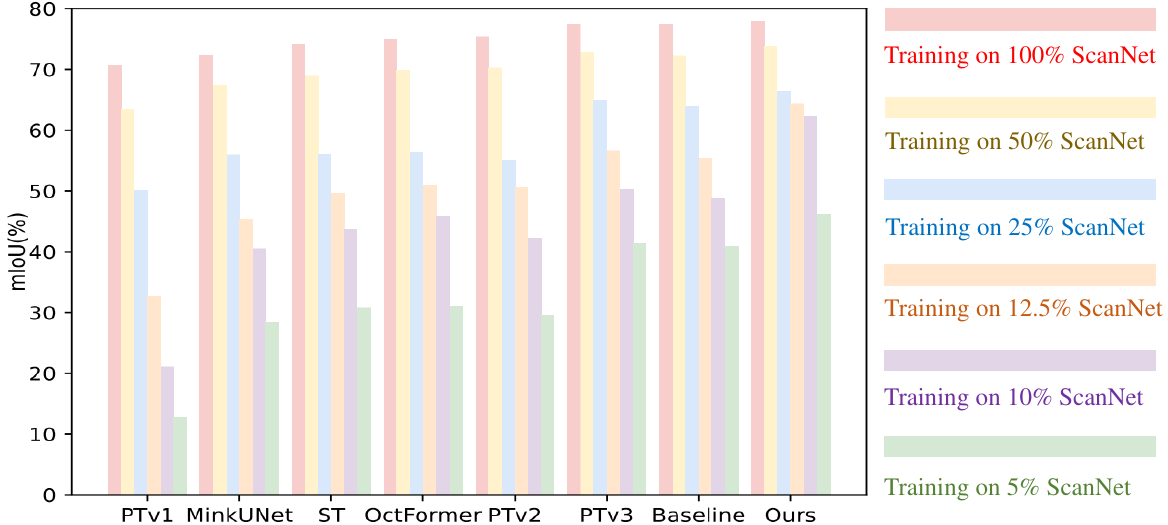}
 \vspace{-0.7cm} 
	\caption{The sparsity robustness results on under-sampled ScanNet. Our method demonstrates the best results in all cases.}
	\label{fig_overfitting_resistance}
\end{figure}
\vspace{-6pt}

Furthermore, we investigate the robustness to noise from other distributions. Tab.~\ref{tab531} shows that although CDSegNet maintains strong performance across noise from other distributions, the performance is slightly reduced compared to Gaussian noise. Combined with Eq.~\ref{f312}, \cite{qu2024conditional} provides an intuitive explanation from the perspective of the gradient of the data distribution, i.e., the predicted noise guiding $\bm{x_t} \stackrel{\bm{\epsilon}}{\longrightarrow} \bm{x_{t-1}}$, $\bm{\epsilon} \sim p_{noise}(\bm{\epsilon}|\bm{x_t},C)$. In this paper, we provide a simpler explanation that is consistent with the source.

According to Sec.~\ref{3.2}, since DDPMs can see multi-level noise samples and fit the noise target from the modeled distribution during training, this makes that they can adapt to related distribution noise during inference. That is, DDPMs are definitely robust to noise from the modeled distribution compared to non-DDPMs. Moreover, the closer the noise distribution is to the modeled distribution, the better the noise robustness (the Laplace distribution); otherwise, this deteriorates (the Poisson distribution).

\vspace{-5pt}
\begin{table}[h]
        \scriptsize
  \resizebox{0.48\textwidth}{!}{
	\begin{tabular}{p{1.2cm}p{0.8cm}p{0.8cm}p{0.7cm}p{0.005cm}p{0.7cm}p{0.7cm}p{0.7cm}}	
        \Xhline{1pt}
  
        \multirow{2}{*}{Methods}
        &\multicolumn{3}{c}{Smalle $\tau$ (mIoU)} 
        &\quad
        &\multicolumn{3}{c}{Big $\tau$ (mIoU) } \\
         \cline{2-4} \cline{6-8}
        
        &\makecell[c]{$\bm{\tau}$=0.01}
        &\makecell[c]{$\bm{\tau}$=0.05}
        &\makecell[c]{$\bm{\tau}$=0.1}
        &\quad
        &\makecell[c]{$\bm{\tau}$=0.5}
        &\makecell[c]{$\bm{\tau}$=0.7}
        &\makecell[c]{$\bm{\tau}$=1.0}\\
        
        \Xhline{1pt}
         
		&\multicolumn{7}{c}{Gaussian Noise} \\
        \cline{2-8}

        PTv2 \cite{wu2022point}
        &\makecell[c]{75.5}
        &\makecell[c]{75.4}
        &\makecell[c]{73.7}
        &\quad
        &\makecell[c]{8.7}
        &\makecell[c]{1.5}
        &\makecell[c]{1.2}\\

        PTv3 \cite{wu2024point}
        &\makecell[c]{\underline{77.6}}
        &\makecell[c]{\underline{77.4}}
        &\makecell[c]{\underline{76.9}}
        &\quad
        &\makecell[c]{\underline{45.8}}
        &\makecell[c]{\underline{26.0}}
        &\makecell[c]{\underline{12.9}}\\

        \rowcolor{gray!20} 
        Ours
        &\makecell[c]{\textbf{77.9}}
       &\makecell[c]{\textbf{77.7}}
        &\makecell[c]{\textbf{77.2}}
        &\quad
        &\makecell[c]{\textbf{57.0}}
        &\makecell[c]{\textbf{46.7}}
        &\makecell[c]{\textbf{35.9}}\\
        \Xhline{1pt}

		&\multicolumn{7}{c}{Uniform Noise} \\
        \cline{2-8}

        PTv2 \cite{wu2022point}
        &\makecell[c]{75.5}
        &\makecell[c]{75.5}
        &\makecell[c]{75.2}
        &\quad
        &\makecell[c]{51.2}
        &\makecell[c]{45.7}
        &\makecell[c]{20.6}\\

        PTv3 \cite{wu2024point}
        &\makecell[c]{\underline{77.6}}
        &\makecell[c]{\underline{77.6}}
        &\makecell[c]{\underline{77.5}}
        &\quad
        &\makecell[c]{\underline{74.3}}
        &\makecell[c]{\underline{70.6}}
       &\makecell[c]{\underline{56.5}}\\

      \rowcolor{gray!20} 
        Ours
        &\makecell[c]{\textbf{77.9}}
        &\makecell[c]{\textbf{77.9}}
        &\makecell[c]{\textbf{77.8}}
        &\quad
        &\makecell[c]{\textbf{74.8}}
        &\makecell[c]{\textbf{70.9}}
        &\makecell[c]{\textbf{56.8}}\\
        \Xhline{1pt}

		&\multicolumn{7}{c}{Laplace Noise} \\
        \cline{2-8}

        PTv2 \cite{wu2022point}
        &\makecell[c]{75.4}
        &\makecell[c]{75.2}
        &\makecell[c]{73.9}
        &\quad
        &\makecell[c]{22.4}
        &\makecell[c]{7.4}
        &\makecell[c]{3.2}\\

        PTv3 \cite{wu2024point}
        &\makecell[c]{\underline{77.6}}
        &\makecell[c]{\underline{77.4}}
        &\makecell[c]{\underline{76.2}}
        &\quad
        &\makecell[c]{\underline{30.1}}
        &\makecell[c]{\underline{14.5}}
        &\makecell[c]{\underline{8.3}}\\

        \rowcolor{gray!20} 
        Ours
        &\makecell[c]{\textbf{77.8}}
        &\makecell[c]{\textbf{77.6}}
        &\makecell[c]{\textbf{76.7}}
        &\quad
        &\makecell[c]{\textbf{43.0}}
        &\makecell[c]{\textbf{26.7}}
        &\makecell[c]{\textbf{14.3}}\\
        \Xhline{1pt}

        &\multicolumn{7}{c}{Possion Noise} \\
        \cline{2-8}

        PTv2 \cite{wu2022point}
        &\makecell[c]{75.5}
        &\makecell[c]{74.2}
        &\makecell[c]{61.2}
        &\quad
        &\makecell[c]{2.4}
        &\makecell[c]{1.2}
        &\makecell[c]{1.0}\\

        PTv3 \cite{wu2024point}
        &\makecell[c]{\underline{77.6}}
        &\makecell[c]{\textbf{77.1}}
        &\makecell[c]{\textbf{73.6}}
        &\quad
        &\makecell[c]{\underline{5.7}}
        &\makecell[c]{\underline{2.8}}
        &\makecell[c]{\underline{1.5}}\\

        \rowcolor{gray!20} 
        Ours
        &\makecell[c]{\textbf{77.8}}
        &\makecell[c]{\underline{76.8}}
        &\makecell[c]{\underline{73.3}}
        &\quad
        &\makecell[c]{\textbf{6.5}}
        &\makecell[c]{\textbf{3.0}}
        &\makecell[c]{\textbf{1.9}}\\
        \Xhline{1pt}
        
	\end{tabular}
 }
	\caption{The results of multiple distribution noise robustness on ScanNet. Our method performs well on the Gaussian and Laplace noise (approximating the modeled distributions), but performs sightly poorly on the Poisson noise (far from the one).}
	\label{tab531}

\end{table}
\vspace{-18pt}

\subsection{Validation for Sparsity Robustness}
\label{5.4}

We also conduct sparsity robustness experiments on the under-sampled ScanNet. This first randomly samples $5\%$, $10\%$, $12.5\%$, $25\%$, and $50\%$ from the training and validation set, respectively. Subsequently, the model is trained and fitted on the under-sampled training and validation set, while performing inference on the entire validation set. 

Fig.~\ref{fig_overfitting_resistance} illustrates the results. CDSegNet demonstrates the significant superiority on under-sampled data. As mentioned in Sec.~\ref{3.2}, CNF aligns a learnable feature augmentation, mapping the features to more complex distributions, enhancing the overfitting resistance of models on less data.


\subsection{Generalization for CNF} 
\label{5.5}

As mentioned in Sec.~\ref{4.1}, CNF is a new network framework that introduces DDPMs into 3D tasks. Therefore, we further conduct the generalization experiments of CNF. 

\textbf{Other backbones.} We first conduct the experiments for introducing CNF into MinkUNet and PTv3. We only add FFM and NN of CDSegNet to MinkUNet (using MinkUNet achieving NN) and PTv3. In Tab.~\ref{tab551}, by introducing CNF, MinkUNet and PTv3 exhibit the better results in multiple benchmarks, due to the feature augmentation through reasonable perturbations. Moreover, benefiting from the lightweight NN, the additional inference time and memory consumption introduced by CNF in MinkUnet and PTv3 are negligible. Furthermore, we observe that PTv3 is surprisingly slower than CDSegNet on nuScenes, as PTv3 utilizes time-consuming point-level element-wise addition for skip connections, while CDSegNet uses channel concatenation.

\vspace{-5pt}
\begin{table}[h]
        \scriptsize
        \resizebox{0.48\textwidth}{!}{
	\begin{tabular}{p{1.7cm}p{0.9cm}p{0.7cm}p{0.7cm}p{0.7cm}p{0.7cm}p{1.1cm}}	

        \Xhline{1pt}
        \makecell[l]{Methods}
        &\makecell[c]{Params}
        &\makecell[c]{mIoU}
        &\makecell[c]{$\tau$=0.1}
        &\makecell[c]{$\tau$=0.5}
        &\makecell[c]{$\tau$=1.0}
        &\makecell[c]{IT/MM}\\
        \hline

	&\multicolumn{6}{c}{ScanNet \cite{dai2017scannet}}\\
        \cline{2-7}

        \makecell[l]{MinkUNet \cite{choy20194d}}
        &\makecell[c]{37.9M}
        &\makecell[c]{72.3}
        &\makecell[c]{70.0}
        &\makecell[c]{3.6}
        &\makecell[c]{0.1}
        &\makecell[c]{63s/0.9G}\\
        
        \makecell[l]{PTv3 \cite{wu2024point}}
        &\makecell[c]{46.2M}
        &\makecell[c]{\underline{77.6}}
        &\makecell[c]{\underline{77.5}}
        &\makecell[c]{45.8}
        &\makecell[c]{12.9}
        &\makecell[c]{53s/1.1G}\\

        \makecell[l]{Baseline}
        &\makecell[c]{101.4M}
        &\makecell[c]{77.5}
        &\makecell[c]{76.9}
        &\makecell[c]{46.1}
        &\makecell[c]{13.0}
        &\makecell[c]{55s/1.9G}\\

        \rowcolor{gray!20} 
        \makecell[l]{MinkUNet+CNF}
        &\makecell[c]{47.1M}
        &\makecell[c]{73.5}
        &\makecell[c]{73.4}
        &\makecell[c]{35.4}
        &\makecell[c]{12.2}
        &\makecell[c]{63s/1.0G}\\

        \rowcolor{gray!20} 
        \makecell[l]{PTv3+CNF}
        &\makecell[c]{59.4M}
        &\makecell[c]{\underline{77.7}}
        &\makecell[c]{77.4}
        &\makecell[c]{\textbf{60.1}}
        &\makecell[c]{\underline{33.2}}
        &\makecell[c]{53s/1.1G}\\

        \rowcolor{gray!20} 
        \makecell[l]{CDSegNet}
        &\makecell[c]{101.4M}
        &\makecell[c]{\textbf{77.9}}
        &\makecell[c]{\textbf{77.6}}
        &\makecell[c]{\underline{57.0}}
        &\makecell[c]{\textbf{35.9}}
        &\makecell[c]{56s/1.9G}\\
        
        \Xhline{1pt}

	&\multicolumn{6}{c}{ScanNet200 \cite{rozenberszki2022language}}\\
        \cline{2-7}

        \makecell[l]{MinkUNet \cite{choy20194d}}
        &\makecell[c]{37.9M}
        &\makecell[c]{28.3}
        &\makecell[c]{27.8}
        &\makecell[c]{1.2}
        &\makecell[c]{0.0}
        &\makecell[c]{64s/0.9G}\\

        \makecell[l]{PTv3 \cite{wu2024point}}
        &\makecell[c]{46.2M}
        &\makecell[c]{35.3}
        &\makecell[c]{34.0}
        &\makecell[c]{10.2}
        &\makecell[c]{1.0}
        &\makecell[c]{53s/1.1G}\\

        \makecell[l]{Baseline}
        &\makecell[c]{101.4M}
        &\makecell[c]{35.4}
        &\makecell[c]{33.9}
        &\makecell[c]{11.1}
        &\makecell[c]{1.2}
        &\makecell[c]{56s/1.9G}\\

        \rowcolor{gray!20} 
        \makecell[l]{MinkUNet+CNF}
        &\makecell[c]{47.1M}
        &\makecell[c]{30.6}
        &\makecell[c]{29.4}
        &\makecell[c]{15.6}
        &\makecell[c]{\underline{6.4}}
        &\makecell[c]{64s/0.9G}\\

        \rowcolor{gray!20} 
        \makecell[l]{PTv3+CNF}
        &\makecell[c]{59.4M}
        &\makecell[c]{\underline{35.9}}
        &\makecell[c]{\underline{35.5}}
        &\makecell[c]{\textbf{21.2}}
        &\makecell[c]{\textbf{7.0}}
        &\makecell[c]{53s/1.1G}\\

        \rowcolor{gray!20} 
        \makecell[l]{CDSegNet}
        &\makecell[c]{101.4M}
        &\makecell[c]{\textbf{36.3}}
        &\makecell[c]{\textbf{35.7}}
        &\makecell[c]{\underline{20.6}}
        &\makecell[c]{5.5}
        &\makecell[c]{57s/1.9G}\\

        \Xhline{1pt}      

	&\multicolumn{6}{c}{nuScenes \cite{caesar2020nuscenes}}\\
        \cline{2-7}

        \makecell[l]{MinkUNet \cite{choy20194d}}
        &\makecell[c]{37.9M}
        &\makecell[c]{73.4}
        &\makecell[c]{45.4}
        &\makecell[c]{1.0}
        &\makecell[c]{1.0}
        &\makecell[c]{733s/1.0G}\\
        
        \makecell[l]{PTv3 \cite{wu2024point}}
        &\makecell[c]{46.2M}
        &\makecell[c]{80.3}
        &\makecell[c]{63.9}
        &\makecell[c]{1.1}
        &\makecell[c]{1.1}
        &\makecell[c]{163s/1.0G}\\

        \makecell[l]{Baseline}
        &\makecell[c]{101.4M}
        &\makecell[c]{80.4}
        &\makecell[c]{64.5}
        &\makecell[c]{1.1}
        &\makecell[c]{1.1}
        &\makecell[c]{128s/1.8G}\\

        \rowcolor{gray!20} 
        \makecell[l]{MinkUNet+CNF}
        &\makecell[c]{47.1M}
        &\makecell[c]{75.7}
        &\makecell[c]{55.3}
        &\makecell[c]{\underline{1.3}}
        &\makecell[c]{\underline{1.3}}
        &\makecell[c]{735s/1.1G}\\

        \rowcolor{gray!20} 
        \makecell[l]{PTv3+CNF}
        &\makecell[c]{59.4M}
        &\makecell[c]{\underline{81.0}}
        &\makecell[c]{\underline{64.8}}
        &\makecell[c]{\underline{1.3}}
        &\makecell[c]{\underline{1.3}}
        &\makecell[c]{126s/1.1G}\\

        \rowcolor{gray!20} 
        \makecell[l]{CDSegNet}
        &\makecell[c]{101.4M}
        &\makecell[c]{\textbf{81.2}}
        &\makecell[c]{\textbf{66.2}}
        &\makecell[c]{\textbf{3.8}}
        &\makecell[c]{\textbf{3.8}}
        &\makecell[c]{129s/1.8G}\\

        \Xhline{1pt}

	\end{tabular}
 }
  \vspace{-0.1cm} 
	\caption{The results of backbones on multiple benchmarks. 'IT' and 'MM' mean the inference time and the mean memory. We run on an  NVIDIA 3090 GPU with batch size=1, without using test-time augmentation (TTA) and fragmented inference.}
	\label{tab551}
\end{table}
\vspace{-15pt}
\vspace{-10pt}
\begin{table}[h]
        \scriptsize
        \resizebox{0.48\textwidth}{!}{
	\begin{tabular}{p{1.7cm}p{0.9cm}p{0.7cm} p{0.7cm} p{2.5cm}p{1.1cm}}	

        \Xhline{1pt}
        \makecell[l]{Methods}
        &\makecell[c]{Params}
        &\makecell[c]{Val}
        &\makecell[c]{Test}
        &\makecell[c]{Only using training data?}
        &\makecell[c]{IT/MM}\\
        \hline

        \makecell[l]{PTv3 \cite{wu2024point}}
        &\makecell[c]{46.2M}
        &\makecell[c]{80.3}
        &\makecell[c]{81.2}
        &\makecell[c]{\Checkmark}
        &\makecell[c]{163s/1.0G}\\
        
        \makecell[l]{PTv3+PPT \cite{wu2024point}}
        &\makecell[c]{97.3M}
        &\makecell[c]{\textbf{81.2}}
        &\makecell[c]{\textbf{83.0}}
        &\makecell[c]{\XSolidBrush}
        &\makecell[c]{-/-}\\

        \rowcolor{gray!20} 
        \makecell[l]{PTv3+CNF}
        &\makecell[c]{59.4M}
        &\makecell[c]{\underline{81.0}}
        &\makecell[c]{\underline{82.8}}
        &\makecell[c]{\Checkmark}
        &\makecell[c]{126s/1.1G}\\

        \rowcolor{gray!20} 
        \makecell[l]{CDSegNet}
        &\makecell[c]{101.4M}
        &\makecell[c]{\textbf{81.2}}
        &\makecell[c]{82.0}
        &\makecell[c]{\Checkmark}
        &\makecell[c]{129s/1.9G}\\
        \Xhline{1pt}

	\end{tabular}
 }
 \vspace{-0.1cm}
	\caption{The results on the nuScenes test set. Our method shows excellent generalization ability on the nuScenes test set. }
	\label{tab553}
\end{table}
\vspace{-10pt}

\textbf{On Test set.} We further evaluated on the nuScenes test set. PTv3+PPT adopts a multi-dataset joint training strategy (For a fair comparison, we avoid deliberately optimizing the results, as we cannot know how many datasets and tricks were actually used). In Fig.~\ref{tab553}, PTv3+CNF, despite having nearly half the number of parameters and being trained only on the training set, performs slightly worse than PTv3+PPT.



\textbf{Other 3D tasks.} We also introduce CNF into the other 3D tasks, classification. PointNet \cite{qi2017pointnet} and PointNet++ \cite{qi2017pointnet++} are selected as backbones. We use an additional PointNet++ branch for modeling the diffusion process.  Tab.~\ref{tab554} show that by introducing CNF, they exhibit significant improvements in classification, alleviating the sensitivity to noise.

\vspace{-10pt}
\begin{table}[h]
        \scriptsize
  \resizebox{0.48\textwidth}{!}{
	\begin{tabular}{p{1.4cm}p{0.5cm}p{0.5cm}p{0.8cm}p{0.6cm}p{0.005cm}p{0.5cm}p{0.5cm}p{0.8cm}p{0.6cm}}	
        \Xhline{1pt}
        
        \multirow{2}{*}{Methods}
        &\multicolumn{4}{c}{PointNet \cite{qi2017pointnet}} 
        &\quad
        &\multicolumn{4}{c}{PointNet++ \cite{qi2017pointnet++}}       \\
        \cline{2-5} \cline{7-10}
        
        &\makecell[c]{CA}
        &\makecell[c]{IA}
        &\makecell[c]{$\tau$=0.5}
        &\makecell[c]{$25\%$}
        &\quad
        &\makecell[c]{CA}
        &\makecell[c]{IA}
        &\makecell[c]{$\tau$=0.5}
        &\makecell[c]{$25\%$}         \\
        \cline{2-5} \cline{7-10}

        Without CNF
        &\makecell[c]{{86.9}}
        &\makecell[c]{{90.5}}
        &\makecell[c]{2.4}
        &\makecell[c]{{19.8}}
        &\quad
        &\makecell[c]{{90.4}}
        &\makecell[c]{{92.3}}
        &\makecell[c]{{2.5}}
        &\makecell[c]{{20.0}}         \\

        \rowcolor{gray!20} 
        With CNF
        &\makecell[c]{\textbf{88.1}}
        &\makecell[c]{\textbf{91.9}}
        &\makecell[c]{\textbf{15.6}}
        &\makecell[c]{\textbf{25.3}}
        &\quad
        &\makecell[c]{\textbf{91.3}}
        &\makecell[c]{\textbf{93.2}}
        &\makecell[c]{\textbf{16.1}}
        &\makecell[c]{\textbf{28.7}}   \\

        \Xhline{1pt}
	\end{tabular}
 }
	\caption{The classification results on ModelNet40 \cite{wu20153d}. `CA` means the class accuracy, while `IA` stands for the instance accuracy. Introducing CNF improves the results in classification.}
	\label{tab554}

\end{table}
\vspace{-20pt}

\subsection{Ablation Study}
\vspace{-3pt}

\begin{figure}[htp]
	\centering
	\includegraphics[width=0.48\textwidth]{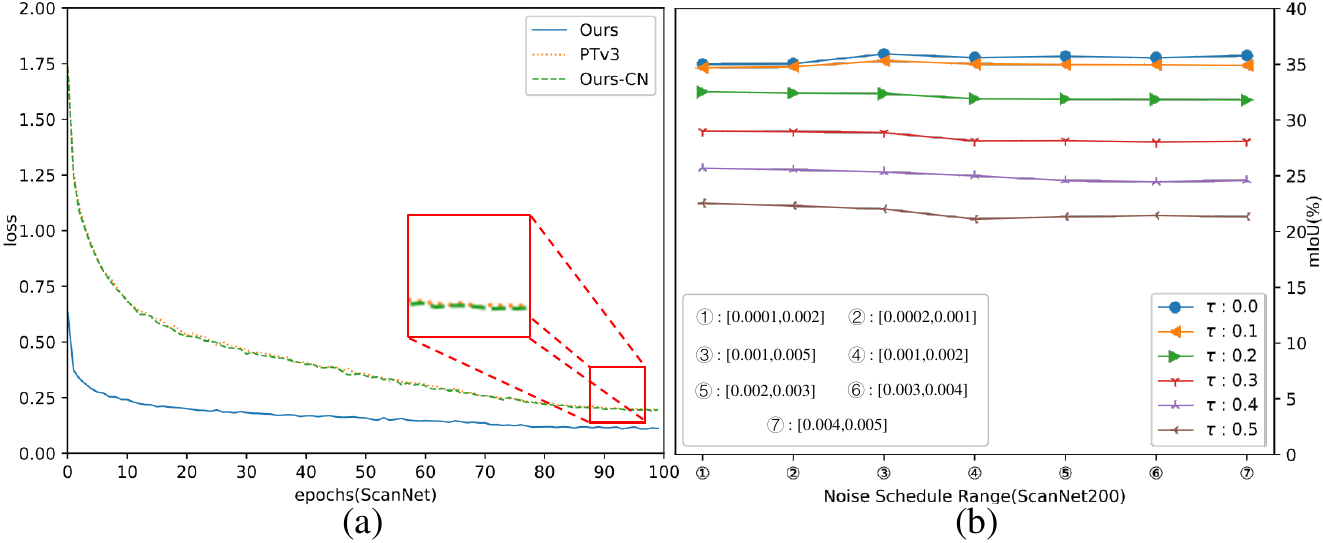}
 \vspace{-0.8cm} 
	\caption{(a) shows the training loss curves of Ours, PTv3 and Ours-CN on ScanNet. The loss curve of Ours-CN (with more parameters compare) is nearly identical to that of PTv3, but the poorer results during inference, indicating overfitting. However, Ours demonstrates the lower loss curve and better performance, as the reasonable perturbations from NN alleviates the model overfitting, enhancing the generalization. In (b), the linear schedule noise range \textcircled{3} demonstrates the best trade-off.}
	\label{fig_loss_range}

\end{figure}

\textbf{The diffusion modeling.} We first conduct the ablation study for the diffusion modeling in CDSegNet. Tab.~\ref{tab561} shows the results on ScanNet.  Ours-CN surprisingly experiences a significant drop across all cases. We believe that the excessive parameters make the model overfit on ScanNet, resulting in the poor generalization. In fact, reducing the parameter number will transform Ours-CN into PTv3. This demonstrates that the reasonable perturbations can significantly enhance the overfitting resistance of CN. Fig.~\ref{fig_loss_range}(a) further supports the conclusion.








\vspace{-8pt}
\vspace{-10pt}

\textbf{The noise schedule range.} The noise schedule range is crucial for CDSegNet, controlling the perturbation degree. Generally, the noise schedule range is positively correlated with the robustness, while showing a negative correlation with the performance. This produces the trade-off between the segmentation performance and the noise robustness.

In Fig.~\ref{fig_loss_range}(b),  \textcircled{3} yields the best trade-off on ScanNet200. Meanwhile, the cosine noise schedule range $[0, 1000]$ and the linear schedule noise range  \textcircled{5} correspond to the best results for ScanNet and nuScenes, respectively. This inspires us to choose \textbf{a smaller schedule range for the more complex (ScanNet200) or sparser (nuScenes) scenes}. Conversely, \textbf{a larger range should be selected (ScanNet)}.

\textbf{The multi-task optimization.} Benefiting from the dual branch architecture \cite{huang2021predator, qu2024conditional}, CDSegNet has two fitting objectives: the  noise fitting and the semantic label fitting, coming from NN and CN, respectively. However, as mentioned in Sec.~\ref{3.3}, the noise fitting typically converges more slowly, due to more fitting targets. This may cause the unreasonable noise information from NN to impair the ability of CN for understanding the scene semantics. Therefore, in addition to FFM, we also apply an effective loss balancing strategy to further alleviates the unreasonable noise perturbations  (more optimizations in the supplementary material). 

We consider several loss balancing strategies: 1) Equal Weighting (EW). 2) Random Loss Weighting (RLW) \cite{lin2021reasonable}. 3) Uncertainty Weights (UW) \cite{kendall2018multi}. 4) Geometric Loss Strategy (GLS) \cite{chennupati2019multinet++}. Tab.~\ref{tab563} shows that GLS produces the best results. Moreover, we observed that after balancing with GLS, the loss values from CN and NN become closer and exhibit smaller fluctuations. \textbf{This can further inspire us to optimize the model with a multi-branch framework from a multi-task perspective.} Meanwhile, to achieve better performance, the loss values among multiple tasks should be more compact and stable.

\vspace{-8pt}
\begin{table}[h]
        \scriptsize
  \resizebox{0.48\textwidth}{!}{
	\begin{tabular}{p{1.0cm}p{0.8cm}p{0.8cm}p{0.8cm}p{0.8cm}p{0.005cm}p{1.2cm}p{1.2cm}}	
        \Xhline{1pt}
  
        \multirow{2}{*}{Methods}
        &\multirow{2}{*}{Params}
        &\multicolumn{3}{c}{Performance} 
        &\quad
        &\multicolumn{2}{c}{Robustness}  \\
         \cline{3-5} \cline{7-8}

        &
        &\makecell[c]{mIoU}
        &\makecell[c]{mAcc}
        &\makecell[c]{allAcc}
        &\quad
        &\makecell[c]{$\tau$=0.5}
        &\makecell[c]{$25\%$}\\
       \hline

        PTv3
        &\makecell[c]{46.2M}
        &\makecell[c]{\underline{77.6}}
        &\makecell[c]{85.0}
        &\makecell[c]{\underline{92.0}}
        &\quad
        &\makecell[c]{45.8}
        &\makecell[c]{\underline{64.3}}\\

        Ours-CN
        &\makecell[c]{88.1M}
        &\makecell[c]{76.6}
        &\makecell[c]{84.6}
        &\makecell[c]{91.6}
        &\quad
        &\makecell[c]{43.4}
        &\makecell[c]{61.8}\\

        Baseline
        &\makecell[c]{101.4M}
        &\makecell[c]{77.5}
        &\makecell[c]{\underline{85.1}}
        &\makecell[c]{91.9}
        &\quad
        &\makecell[c]{\underline{46.1}}
        &\makecell[c]{64.0}\\

        \rowcolor{gray!20} 
        Ours
        &\makecell[c]{101.4M}
        &\makecell[c]{\textbf{77.9}}
        &\makecell[c]{\textbf{85.2}}
        &\makecell[c]{\textbf{92.2}}
        &\quad
        &\makecell[c]{\textbf{57.0}}
        &\makecell[c]{\textbf{66.5}}\\

        \Xhline{1pt}
        
	\end{tabular}
 }
	\caption{Ablation study of the diffusion modeling in CDSegNet on ScanNet. The baseline means removing the diffusion modeling in NN. Meanwhile, Our-CN represents removing entire NN in CDSegNet, retaining only CN. This results demonstrate the importance of the diffusion modeling for CDSegNet.}
	\label{tab561}

\end{table}
\vspace{-25pt}

\begin{table}[h]
        \scriptsize
  \resizebox{0.48\textwidth}{!}{
	\begin{tabular}{p{1.0cm}p{0.8cm}p{0.8cm}p{0.8cm}p{0.005cm}p{1.2cm}p{1.2cm}}	
        \Xhline{1pt}
  
        \multirow{2}{*}{Methods}
        &\multicolumn{3}{c}{Performance} 
        &\quad
        &\multicolumn{2}{c}{Robustness} \\
         \cline{2-4} \cline{6-7}

        &\makecell[c]{mIoU}
        &\makecell[c]{mAcc}
        &\makecell[c]{allAcc}
        &\quad
        &\makecell[c]{$\tau$=0.5}
        &\makecell[c]{$25\%$}\\
       \hline

        EW
        &\makecell[c]{\underline{77.6}}
        &\makecell[c]{\underline{85.1}}
        &\makecell[c]{92.0}
        &\quad
        &\makecell[c]{55.5}
        &\makecell[c]{64.4}\\

        RLW \cite{lin2021reasonable}
        &\makecell[c]{77.4}
        &\makecell[c]{85.0}
        &\makecell[c]{91.9}
        &\quad
        &\makecell[c]{54.1}
        &\makecell[c]{63.9}\\

        UW \cite{kendall2018multi}
        &\makecell[c]{\underline{77.6}}
        &\makecell[c]{{85.0}}
        &\makecell[c]{\underline{92.1}}
        &\quad
        &\makecell[c]{\underline{55.6}}
        &\makecell[c]{\underline{65.1}}\\

        \rowcolor{gray!20} 
        GLS \cite{chennupati2019multinet++}
        &\makecell[c]{\textbf{77.9}}
        &\makecell[c]{\textbf{85.2}}
        &\makecell[c]{\textbf{92.2}}
        &\quad
        &\makecell[c]{\textbf{57.0}}
        &\makecell[c]{\textbf{66.5}}\\

        \Xhline{1pt}
        
	\end{tabular}
 }
	\caption{Ablation study of multi-task optimization on ScanNet. GLS shows the better performance and robustness for CDSegNet.}
	\label{tab563}

\end{table}
\vspace{-25pt}

\section{Conclusion}
\vspace{-8pt}
In this paper, we systematically revealed some key insights of DDPMs in point cloud semantic segmentation. Meanwhile, to preserve the advantages while overcoming the limitations, a Conditional-Noise Framework  of DDPMs is designed. Based on this, we further proposed an end-to-end robust point cloud segmentation network, CDSegNet. CDSegNet demonstrated the strong robustness to the data noise and sparsity, while requiring a single-step inference. Moreover, we provided a more understandable explanation for the noise robustness from DDPMs. Overall, we have lowered the barrier for applying DDPMs, with the hope of encouraging broader extensions of DDPMs in 3D tasks.

\textbf{Acknowledgments.} 
This work was supported in part by the Frontier Technologies R$\&$D Program of Jiangsu under grant BF2024070, in part by the National Natural Science Foundation of China under Grant 62471235, in part by Hunan Natural Science Foundation Project (No. 2025JJ50338) and Shanghai Education Committee AI Project (No. JWAIYB-2).


{
    \small
    \bibliographystyle{ieeenat_fullname}
    \bibliography{main}
}

\newpage

\setcounter{equation}{0}
\setcounter{section}{0}
\setcounter{figure}{0}
\setcounter{table}{0}

Due to the space limitation of the main text, we include additional experiments, derivations, implementations, and discussions in the supplementary material. We first conduct additional ablation (Sec.~\ref{sec1}) and comparison (Sec.~\ref{sec2}) experiments. Then, we logically derive DDPMs from a modeling perspective, explaining some issues about applying DDPMs to 3D tasks (Sec.~\ref{sec3}). Next, the implementation details (Sec.~\ref{sec4}) and optimization process (Sec.~\ref{sec5}) of our method are presented. Finally, we discuss the limitations of CNF (Sec.~\ref{sec6}) and visualize additional results (Sec.~\ref{sec7}).

\section{Additional Ablation Study}
\label{sec1}

\subsection{Selection of FFM}

Since excessive noise perturbations from the Noise Network (NN) may harm the performance of the Conditional Network (CN), the aim of the Feature Fusion Module (FFM) is to adaptively filter the noise information, making the feature augmentation in a reasonable way. To better achieve the aim, we consider several ways of FFM: 1) Channel Mapping (CM) \cite{lyu2021conditional}. This preserves the channel information of features from CN and NN, but lacks the effective information filtering in the feature space. 2) Channel Cross-attention (CCA) \cite{zamir2022restormer}. This filters information along the channel dimension, but compresses the search space, making filtering out effective information difficult at the point level. 3) Spatial Cross-Attention (SCA) \cite{vaswani2017attention}. This precisely searches for similar elements in the spatial dimension, but the quadratic complexity for the input points. Fortunately, the input point number at the bottleneck stage of the U-Net is usually less than a thousand.

Tab.~\ref{supp_tab111} exhibits the results on ScanNet. Benefiting from the effective filtering for perturbations, spatial cross-attention significantly outperforms  other two ways. 

\subsection{Inference Modes} 

Although CDSegNet can be considered a non-DDPM during inference, CDSegNet can still follow the iterative inference approach of DDPMs (the output is still dominated by CN). Therefore, this can be divided into three inference modes: 1) Single-Step Inference (SSI), semantic labels are generated by CN through a single-step iteration in NN. 2) Multi-Step Average Inference (MSAI), MSAI conducts  $T$ step iterations in NN and averages $T$ outputs produced by CN. 3) Multi-Step Final Inference (MSFI), MSFI is determined by the output from the final iteration of CN.

Tab.~\ref{supp_tab112} shows that the difference in performance among SSI, MSAI, and MSFI is negligible. Thanks to CNF, the impact of iterations is no longer significant for the results. Meanwhile, since the noise system of DDPMs is retained during training, CDSegNet still maintains the robustness to data noise and sparsity.

\begin{table}[h]
        \scriptsize
  \resizebox{0.48\textwidth}{!}{
	\begin{tabular}{p{1.0cm}p{0.8cm}p{0.8cm}p{0.8cm}p{0.005cm}p{1.2cm}p{1.2cm}}	
       \Xhline{1pt}
  
        \multirow{2}{*}{Methods}
        &\multicolumn{3}{c}{Performance} 
        &\quad
        &\makecell[c]{Training}
        &\makecell[c]{Inference} \\
         \cline{2-4} \cline{6-7}

        &\makecell[c]{mIoU}
        &\makecell[c]{mAcc}
        &\makecell[c]{allAcc}
        &\quad
        &\makecell[c]{Latency}
        &\makecell[c]{Latency}\\
       \hline

        CM \cite{lyu2021conditional}
        &\makecell[c]{77.4}
        &\makecell[c]{84.8}
        &\makecell[c]{91.9}
        &\quad
        &\makecell[c]{\textbf{268ms}}
        &\makecell[c]{\textbf{101ms}}\\

        CCA \cite{zamir2022restormer}
        &\makecell[c]{\underline{77.5}}
        &\makecell[c]{\underline{85.0}}
        &\makecell[c]{\underline{92.3}}
        &\quad
        &\makecell[c]{\underline{272ms}}
        &\makecell[c]{\underline{107ms}}\\

        \rowcolor{gray!20} 
        SCA \cite{vaswani2017attention}
        &\makecell[c]{\textbf{77.9}}
        &\makecell[c]{\textbf{85.2}}
        &\makecell[c]{\textbf{92.2}}
        &\quad
        &\makecell[c]{278ms}
        &\makecell[c]{112ms}\\

        \Xhline{1pt}
        
	\end{tabular}
}
	\caption{Ablation study of FFM on ScanNet. The spatial cross-attention exhibits the most optimal selection.}
	\label{supp_tab111}

\end{table}

\vspace{-10pt}

\begin{table}[h]
        \scriptsize
  \resizebox{0.48\textwidth}{!}{
	\begin{tabular}{p{1.8cm}p{1.4cm}p{1.4cm}p{1.4cm}}	
        \Xhline{1pt}

        Methods
        &\makecell[c]{mIoU}
        &\makecell[c]{mAcc}
        &\makecell[c]{allAcc}\\
       \hline

        MSAI-100
        &\makecell[c]{\textbf{77.9}}
        &\makecell[c]{\textbf{85.3}}
        &\makecell[c]{{92.1}}\\

        MSAI-50
        &\makecell[c]{\underline{77.8}}
        &\makecell[c]{{85.1}}
        &\makecell[c]{{91.9}}\\

        MSAI-20
        &\makecell[c]{\underline{77.8}}
        &\makecell[c]{{85.1}}
        &\makecell[c]{{91.9}}\\

        \hline
        
        MSFI-100
        &\makecell[c]{\underline{77.8}}
        &\makecell[c]{\underline{85.2}}
        &\makecell[c]{\textbf{92.3}}\\

        MSFI-50
        &\makecell[c]{{77.7}}
        &\makecell[c]{{85.1}}
        &\makecell[c]{{92.1}}\\

        MSFI-20
        &\makecell[c]{{77.7}}
        &\makecell[c]{{85.1}}
        &\makecell[c]{{92.1}}\\

        \hline

        \rowcolor{gray!20}
        SSI
        &\makecell[c]{\textbf{77.9}}
        &\makecell[c]{\underline{85.2}}
        &\makecell[c]{\underline{92.2}}\\

        \Xhline{1pt}
        
	\end{tabular}
 }
	\caption{Ablation study of different inference modes on ScanNet. SSI demonstrates a better trade-off between performance and efficiency.}
	\label{supp_tab112}
\end{table}

\vspace{-10pt}

\begin{table}[h]
        \scriptsize
  \resizebox{0.48\textwidth}{!}{
	\begin{tabular}{p{1.8cm}p{1.4cm}p{1.4cm}p{1.4cm}}	
        \Xhline{1pt}

        Input type
        &\makecell[c]{mIoU}
        &\makecell[c]{mAcc}
        &\makecell[c]{allAcc}\\
       \hline

        Semantic Label
        &\makecell[c]{\underline{77.7}}
        &\makecell[c]{\underline{85.1}}
        &\makecell[c]{\textbf{92.2}}\\

        Point Coordinate
        &\makecell[c]{77.6}
        &\makecell[c]{85.0}
        &\makecell[c]{\underline{92.1}}\\

        \rowcolor{gray!20}
        Color+Normal
        &\makecell[c]{\textbf{77.9}}
        &\makecell[c]{\textbf{85.2}}
        &\makecell[c]{\textbf{92.2}}\\

        \Xhline{1pt}
        
	\end{tabular}
 }
	\caption{Ablation study of input for NN on ScanNet. Color+Normal, consistent with the input of CN, demonstrates the best performance.}
	\label{supp_tab113}
\end{table}

\subsection{Input for NN}

According to Sec.~4.1 of the main text, NN is modeled as a noise-feature generator to enhance the semantic features in CN. Nevertheless, we can still input semantic labels or point coordinates into NN for the diffusion modeling (the color and the normal as the inputs of CN). 

Tab.~\ref{supp_tab113} shows that using the color and the normal as the inputs of NN, consistent with the input of CN, achieves the best results. This demonstrates that the feature perturbations from NN can effectively enhance the semantic features in CN.

\begin{table}[h]
        \scriptsize
  \resizebox{0.48\textwidth}{!}{
	\begin{tabular}{p{1.4cm}p{0.6cm}p{0.6cm}p{0.7cm}p{0.005cm}p{0.7cm}p{0.7cm}p{0.7cm}}	
        \Xhline{1pt}
  
        \multirow{2}{*}{Methods}
        &\multicolumn{3}{c}{Performance} 
        &\quad
        &\multicolumn{3}{c}{Robustness} \\
         \cline{2-4} \cline{6-8}
        
        &\makecell[c]{mIoU}
        &\makecell[c]{mAcc}
        &\makecell[c]{allAcc}
        &\quad
        &\makecell[c]{$\bm{\tau}$=0.1}
        &\makecell[c]{$\bm{\tau}$=0.5}
        &\makecell[c]{$\bm{\tau}$=1.0}\\
        
        \Xhline{1pt}
         
		&\multicolumn{7}{c}{ScanNet \cite{dai2017scannet}} \\
       \cline{2-8}

       PTv3 \cite{wu2024point}
        &\makecell[c]{77.6}
        &\makecell[c]{\textbf{85.0}}
        &\makecell[c]{\textbf{92.0}}
        &\quad
        &\makecell[c]{\textbf{77.5}}
        &\makecell[c]{45.8}
        &\makecell[c]{12.9}\\

        \rowcolor{gray!20} 
        PTv3 + CNF
        &\makecell[c]{\textbf{77.7}}
        &\makecell[c]{84.8}
        &\makecell[c]{91.7}
        &\quad
        &\makecell[c]{77.4}
        &\makecell[c]{\textbf{60.1}}
        &\makecell[c]{\textbf{33.2}}\\
        \Xhline{1pt}

	&\multicolumn{7}{c}{ScanNet200 \cite{rozenberszki2022language}} \\
        \cline{2-8}

        PTv3 \cite{wu2024point}
        &\makecell[c]{35.3}
        &\makecell[c]{\textbf{46.0}}
        &\makecell[c]{83.3}
        &\quad
        &\makecell[c]{34.0}
        &\makecell[c]{10.2}
        &\makecell[c]{1.0}\\

        \rowcolor{gray!20} 
        PTv3 + CNF
        &\makecell[c]{\textbf{35.9}}
        &\makecell[c]{45.3}
        &\makecell[c]{\textbf{83.4}}
        &\quad
        &\makecell[c]{\textbf{35.5}}
        &\makecell[c]{\textbf{21.2}}
        &\makecell[c]{\textbf{7.1}}\\
        \Xhline{1pt}

	&\multicolumn{7}{c}{nuScenes \cite{caesar2020nuscenes}} \\
        \cline{2-8}

        PTv3 \cite{wu2024point}
        &\makecell[c]{{80.3}}
        &\makecell[c]{{87.2}}
        &\makecell[c]{{94.6}}
        &\quad
        &\makecell[c]{{63.9}}
        &\makecell[c]{{1.1}}
        &\makecell[c]{{1.1}}\\

        \rowcolor{gray!20} 
        PTv3 + CNF
        &\makecell[c]{\textbf{81.0}}
        &\makecell[c]{\textbf{87.9}}
        &\makecell[c]{\textbf{94.8}}
        &\quad
        &\makecell[c]{\textbf{67.8}}
        &\makecell[c]{\textbf{1.3}}
        &\makecell[c]{\textbf{1.3}}\\
        \Xhline{1pt}
        
	\end{tabular}
 }
	\caption{The results of introducing CNF to PTv3. PTv3+CNF shows a significant improvement in noise robustness.}
	\label{supp_tab211}

\end{table}
\vspace{-10pt}

\begin{figure*}[htp]
	\centering
	\includegraphics[width=\textwidth]
 {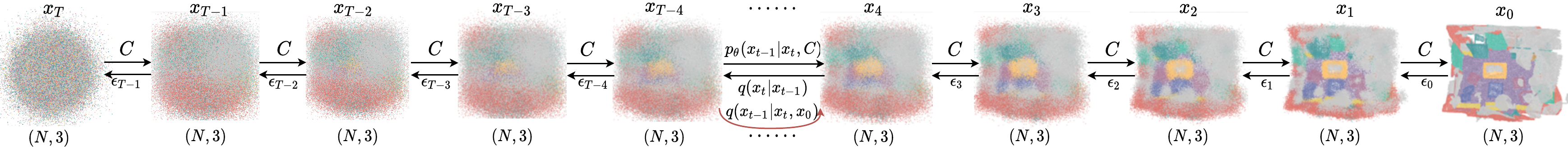}
 \vspace{-0.6cm} 
	\caption{The visualization of the predefined diffusion process $q(\bm{x_t}|\bm{x_{t-1}})$, the  inverse of the diffusion process $q(\bm{x_{t-1}}|\bm{x_t},\bm{x_0})$, and the trainable conditional generation process $p_\theta(\bm{x_{t-1}}|\bm{x_t},C)$. In the diffusion process $q(\bm{x_t}|\bm{x_{t-1}})$, the task target $\bm{x_0}$ is gradually noised until $\bm{x_0}$ degrades to $\bm{z}$ ($\bm{x_T}$). Meanwhile, the inverse of the diffusion process (the true Ground Truth in DDPMs) $q(\bm{x_{t-1}}|\bm{x_t},\bm{x_0})$ can be calculated by the predefined distribution in the diffusion process. Furthermore, the generation process $p_\theta(\bm{x_{t-1}}|\bm{x_t},C)$ gradually fits the inverse of the diffusion process $q(\bm{x_{t-1}}|\bm{x_t},\bm{x_0})$ until $\bm{z}$ ($\bm{x_T}$) is restored to $\bm{x_0}$ conditioned on $C=\{\bm{c},t\}$ (unconditional generation, $\bm{c}=\emptyset$, in the formula derivation of DDPMs, the time label $t$ is usually ignored.).}
	\label{fig_supp_ddpm}
\end{figure*}
\vspace{-10pt}

\section{Additional Comparative Experiments}
\label{sec2}

\subsection{Generalization for CNF on Indoor Benchmark}

We also conduct the generalization experiments of CNF on indoor benchmarks (ScanNet \cite{dai2017scannet}, ScanNet200 \cite{rozenberszki2022language}). Following the same setup as in Sec.~5.5 of the main text, we simply consider PTv3 as CN and add NN and FFM of CDSegNet to PTv3.

Tab.~\ref{supp_tab211} shows this result. By introducing CNF, PTv3 has significantly improved noise robustness. Simultaneously, we can see that although the noise robustness has increased significantly on ScanNet, the performance has only slightly improved. This is because ScanNet represents a purer and denser scene compared to ScanNet200 and nuScenes. This verifies that CNF is more suitable for disturbed and sparse scenes but is slightly inferior in relatively pure scenes.

\subsection{Comparison on Test Set}

We further provide the results on the test sets of ScanNet, ScanNet200 and nuScenes. Sincerely and honestly, all models are uniformly trained on the training set and validated on the validation set. The all comparison methods come from the official released checkpoints. Due to the limitations of the submission number and the checkpoint releases, we provided only a small set of results (\textbf{all results on the test set are from previous checkpoints}).

\textbf{On ScanNet and ScanNet200 test set.} Tab.~\ref{supp_tab221} shows the results of our method compared with PonderV2 \cite{zhu2023ponderv2} and PTv3 \cite{wu2024point} on the test sets of ScanNet and ScanNet200. We can clearly observe that, compared to the results on the validation set, our method performs better on the test set. This demonstrates that reasonable noise perturbations can effectively enhance the generalization ability of models.

\textbf{On nuScenes test set.} As shown in Tab.~\ref{supp_tab222}, CDSegNet significantly outperforms PTv3 on both the validation and test sets. Notably, we found that PTv3 with CNF introduced achieves a significant improvement on the test set, even surpassing CDSegNet. This further demonstrates that CNF enables  models to inherit the sparse robustness from DDPMs, enhancing the generalization ability of models in sparse scenes (we sincerely reaffirm that we only trained on the training set and validated on the validation set. Meanwhile, all checkpoints are downloaded directly from the official website).

\begin{table}[h]
        \scriptsize
  \resizebox{0.48\textwidth}{!}{
	\begin{tabular}{p{1.5cm}p{1.2cm}p{1.3cm}p{2.0cm}p{1.2cm}}	
        \Xhline{1pt}

        Methods
        &\makecell[c]{Val (mIoU)}
        &\makecell[c]{Test (mIoU)}
        &\makecell[c]{Only Training Set?}
        &\makecell[c]{URL}\\
       \Xhline{1pt}

        &\multicolumn{4}{c}{ScanNet \cite{dai2017scannet}} \\
        \cline{2-5}

        PonderV2 \cite{zhu2023ponderv2}
        &\makecell[c]{{77.0}}
        &\makecell[c]{\underline{73.9}}
        &\makecell[c]{no}
        &\makecell[c]{\href{https://github.com/OpenGVLab/PonderV2/blob/main/docs/model_zoo.md}{Here}}\\

        \cline{1-5}
        
        PTv3 \cite{wu2024point}
        &\makecell[c]{\underline{77.6}}
        &\makecell[c]{{73.6}}
        &\makecell[c]{yes}
        &\makecell[c]{\href{https://huggingface.co/Pointcept/PointTransformerV3/tree/main/scannet-semseg-pt-v3m1-0-base/model}{Here}}\\

        \rowcolor{gray!20}
        Ours
        &\makecell[c]{\textbf{77.9}}
        &\makecell[c]{\textbf{74.5}}
        &\makecell[c]{yes}
        &\makecell[c]{-}\\
        \Xhline{1pt}

        &\multicolumn{4}{c}{ScanNet200 \cite{rozenberszki2022language}} \\
        \cline{2-5}

        PTv3 \cite{wu2024point}
        &\makecell[c]{{35.3}}
        &\makecell[c]{{33.2}}
        &\makecell[c]{yes}
        &\makecell[c]{\href{https://huggingface.co/Pointcept/PointTransformerV3/tree/main/scannet200-semseg-pt-v3m1-0-base/model}{Here}}\\
        \hline

        \rowcolor{gray!20}
        PTv3+CNF
        &\makecell[c]{\underline{35.5}}
        &\makecell[c]{\underline{33.7}}
        &\makecell[c]{yes}
        &\makecell[c]{-}\\

        \rowcolor{gray!20}
        Ours
        &\makecell[c]{\textbf{36.0}}
        &\makecell[c]{\textbf{34.1}}
        &\makecell[c]{yes}
        &\makecell[c]{-}\\

        \Xhline{1pt}
	\end{tabular}
 }
	\caption{The results on the ScanNet and ScanNet200 test set. Our method demonstrates better performance on the validation set and the test set.}
	\label{supp_tab221}
\end{table}

\begin{table}[h]
        \scriptsize
  \resizebox{0.48\textwidth}{!}{
	\begin{tabular}{p{1.4cm}p{1.2cm}p{1.3cm}p{2.0cm}p{1.4cm}}	
        \Xhline{1pt}

        Methods
        &\makecell[c]{Val (mIoU)}
        &\makecell[c]{Test (mIoU)}
        &\makecell[c]{Only Training Set?}
        &\makecell[c]{URL}\\
       \Xhline{1pt}

        PTv3 \cite{wu2024point}
        &\makecell[c]{{80.3}}
        &\makecell[c]{{81.2}}
        &\makecell[c]{yes}
        &\makecell[c]{\href{https://huggingface.co/Pointcept/PointTransformerV3/tree/main/nuscenes-semseg-pt-v3m1-0-base/model}{Here}}\\

        \hline
        
        \rowcolor{gray!20}
        PTv3+CNF
        &\makecell[c]{\underline{80.8}}
        &\makecell[c]{\textbf{82.8}}
        &\makecell[c]{yes}
        &\makecell[c]{-}\\

        \rowcolor{gray!20}
        Ours
        &\makecell[c]{\textbf{81.2}}
        &\makecell[c]{\underline{82.0}}
        &\makecell[c]{yes}
        &\makecell[c]{-}\\

        \Xhline{1pt}
	\end{tabular}
 }
	\caption{The results on the nuScenes test set. PTv3+CNF demonstrates a significant improvement compared with PTv3 on the test set.}
	\label{supp_tab222}
\end{table}

\begin{table}[h]
        \scriptsize
  \resizebox{0.48\textwidth}{!}{
	\begin{tabular}{p{1.6cm}p{0.8cm}p{1.2cm}p{1.2cm}p{0.005cm}p{0.8cm}p{0.8cm}}	
        \Xhline{1pt}
        
       \multirow{2}{*}{Methods}
        &\multicolumn{3}{c}{Performance} 
        &\quad
        &\multicolumn{2}{c}{Robustness (mAcc)} \\
        \cline{2-4} \cline{6-7}
        
        &\makecell[c]{mAcc}
        &\makecell[c]{CAmIoU}
        &\makecell[c]{IAmIoU}
        &\quad
        &\makecell[c]{$\tau$=0.5}
        &\makecell[c]{$25\%$}\\
        \Xhline{1pt}

        &\multicolumn{6}{c}{PointNet \cite{qi2017pointnet}} \\
        \cline{2-7}

        Without CNF
        &\makecell[c]{{93.2}}
        &\makecell[c]{{77.9}}
        &\makecell[c]{{83.2}}
        &\quad
        &\makecell[c]{37.3}
        &\makecell[c]{{70.6}}\\        

        \rowcolor{gray!20} 
        With CNF
        &\makecell[c]{\textbf{94.1}}
        &\makecell[c]{\textbf{78.5}}
        &\makecell[c]{\textbf{83.9}}
        &\quad
        &\makecell[c]{\textbf{42.2}}
        &\makecell[c]{\textbf{73.1}}\\  

        \Xhline{1pt}        
        &\multicolumn{6}{c}{PointNet++ \cite{qi2017pointnet++}} \\
        \cline{2-7}

       Without CNF
        &\makecell[c]{{94.2}}
        &\makecell[c]{{82.7}}
        &\makecell[c]{{85.1}}
        &\quad
        &\makecell[c]{{39.0}}
        &\makecell[c]{{71.1}}\\        

        \rowcolor{gray!20} 
        With CNF
        &\makecell[c]{\textbf{95.1}}
        &\makecell[c]{\textbf{83.5}}
        &\makecell[c]{\textbf{86.0}}
        &\quad
        &\makecell[c]{\textbf{44.6}}
        &\makecell[c]{\textbf{73.9}}\\  

        \Xhline{1pt}
	\end{tabular}
 }
	\caption{The instance segmentation results on ShapeNet \cite{chang2015shapenet}. With the introduction of CNF, the performance and robustness of PointNet and PointNet++  have been improved significantly.}
	\label{tab223}

\end{table}

\textbf{Other 3D tasks.}  We also introduce CNF into instance segmentation task. Similar to the classification task in the main text, PointNet \cite{qi2017pointnet} and PointNet++ \cite{qi2017pointnet++} are selected as backbones.  Tab.~\ref{tab223} show that by introducing CNF, they exhibit significantly improve the performance and robustness.

\section{Formula Derivation of DDPMs for 3D Tasks}
\label{sec3}

In this section, we provide the theoretical support for applying DDPMs to most existing 3D vision tasks. This focuses on the logical modeling process of DDPMs, omitting some derivation details that we consider unnecessary. For example, to better understand the modeling process of DDPMs intuitively, we believe that the derivation of the \textbf{E}vidence \textbf{L}ower \textbf{BO}und (ELBO) can be skipped. Interested readers can refer to \cite{qu2024conditional} for the derivation of the ELBO under specific conditions.

For a 3D task, given a data sample pair $(\bm{c}, \bm{x_0})$, we aim to train a generalized model $f_\theta$ that takes $\bm{c}$ as the input and produces an output $\bm{x'_0}$ approximating $\bm{x_0}$. We can transform the task into a conditional generation problem using DDPMs. This performs an auto-regressive process \cite{song2020denoising, qu2024conditional}: a predefined diffusion process $q$ ($\bm{x_0} \rightarrow \bm{z}$) and a trainable conditional generation process $p_\theta$ ($\bm{z} \rightarrow \bm{x_0}$) under the guidance of the condition $\bm{c}$ (see Fig.~\ref{fig_supp_ddpm}). Here, $\bm{z}$ represents an implicit variable sampled from a predefined prior distribution $P_{noise}$ in DDPMs.

\subsection{Diffusion Process}  
\label{sec3.1}

\textbf{The modeled distribution and noise-adding pattern.} The diffusion process $q$ is more critical in DDPMs, as this defines the type of DDPMs. For example, \cite{ho2020denoising} can be referred to as Gaussian or continuous DDPMs. Similarly, we can also use the categorical distribution to model the diffusion process, referred to as categorical or discrete DDPMs \cite{austin2021structured}. Theoretically, any distribution can be used to model the diffusion process, such as the Laplace distribution or the Poisson distribution. Meanwhile, the noise addition pattern is not limited to element-wise addition and multiplication. This can also involve snowification and masking \cite{bansal2024cold}.

We derive this diffusion process based on \cite{ho2020denoising}, which involves adding perturbations using a Gaussian distribution through element-wise addition and multiplication.

\textbf{The detailed derivation.} The diffusion process $q$ is structured as a Markov chain, with each step governed by an independent Gaussian distribution. This gradually destroys the essential information until $\bm{x_0}$ degrades to $\bm{z}$. Meanwhile, this process is only related to the Ground Truth (task target) $\bm{x_0}$ and is independent of the condition (task input) $\bm{c}$. Formally, given a time label $t \sim \mathcal{U}(T)$ that controls the noise level, the diffusion process can be computed by multiple conditional distributions $q(\bm{x_{1:T}}|\bm{x_0})$:

\begin{equation}
\begin{split}
	\label{supp_f311}
q(\bm{x_{1:T}}|\bm{x_0})= \frac{q(\bm{x_{0:T}})}{q(\bm{x_0})} \quad\quad\quad\quad\quad\quad\quad\quad\quad\quad\quad\quad\quad\quad\;\;\\
= \frac{\textcolor{red}{q(\bm{x_T}|\bm{x_{0:T-1}})}q(\bm{x_{0:T-1}})}{q(\bm{x_0})}.\quad\quad\quad\quad\quad\quad\quad\\
\end{split}
\end{equation}

Meanwhile, according to the Markov property, the current term $x_T$ is determined only by the previous term $\bm{x_{T-1}}$: 

\begin{equation}
\begin{split}
	\label{supp_f312}
q(\bm{x_{1:T}}|\bm{x_0})= \frac{\textcolor{red}{q(\bm{x_T}|\bm{x_{T-1}})}q(\bm{x_{T-1}}|\bm{x_{0:T-2}})q(\bm{x_{0:T-2}})}{q(\bm{x_0})}\quad\quad\\ 
= \frac{q(\bm{x_T}|\bm{x_{T-1}})q(\bm{x_{T-1}}|\bm{x_{T-2}})...q(\bm{x_1}|\bm{x_0})\textcolor{red}{\cancel{q(\bm{x_0})}}}{{\textcolor{red}{\cancel{q(\bm{x_0})}}}}\\
= \prod_{t=1}^{T}q(\bm{x_t}|\bm{x_{t-1}}),\quad\quad\quad\quad\quad\quad\quad\quad\quad\quad\quad\quad\;\;
\end{split}
\end{equation}

where $q(\bm{x_t}|\bm{x_{t-1}}) = \mathcal{N}(\bm{x_t}; \sqrt{1-\beta_t}\bm{x_{t-1}},\beta_t\bm{I})$.  $\beta_t$ is a predefined and increasing variance factor. 

\textbf{The intuitive explanation.} We can intuitively understand this diffusion process. According to Fig.~\ref{fig_supp_ddpm}, $\bm{x_1}$ is determined by $\bm{x_0}$, $\bm{x_1} \sim q(\bm{x_1}|\bm{x_0})$. Similarly, $\bm{x_2}$ is determined by both $\bm{x_1}$ and $\bm{x_0}$, $\bm{x_2} \sim q(\bm{x_2}|\bm{x_1},\bm{x_0})$. In this way, $\bm{x_T} \sim q(\bm{x_T}|\bm{x_{0:T-1}})$. Due to the  independent and identically distributed (i.i.d.) and Markov properties, this process can be described as:

\begin{equation}
\begin{split}
	\label{supp_f313}
\textcolor{blue}{i.i.d.:} \quad\quad\quad\quad\quad\quad\quad\quad\quad\quad\quad\quad\quad\quad\quad\quad\quad\quad\quad\quad\quad\quad\quad\quad\quad\quad\quad\quad\quad\quad\quad\quad\;\;\\ 
q(\bm{x_1}|\textcolor{red}{\bm{x_0}})q(\bm{x_2}|\textcolor{red}{\bm{x_1}},\textcolor{red}{\bm{x_0}})...q(\bm{x_T}|\textcolor{red}{\bm{x_{0:T-1}}})\quad\quad\quad\quad\quad\quad\quad\quad\quad\quad\quad\quad \\
\textcolor{blue}{Markov:} \quad\quad\quad\quad\quad\quad\quad\quad\quad\quad\quad\quad\quad\quad\quad\quad\quad\quad\quad\quad\quad\quad\quad\quad\quad\quad\quad\quad\quad\quad\;\;\;\\
=q(\bm{x_1}|\textcolor{red}{\bm{x_0}})q(\bm{x_2}|\textcolor{red}{\bm{x_1}})...q(\bm{x_T}|\textcolor{red}{\bm{x_{T-1}}})\quad\quad\quad\quad\quad\quad\quad\quad\quad\quad\quad\quad\quad\quad\;\\
= \prod_{t=1}^{T}q(\bm{x_t}|\bm{x_{t-1}}).\quad\quad\quad\quad\quad\quad\quad\quad\quad\quad\quad\quad\quad\quad\quad\quad\quad\quad\quad\quad\quad\;\;
\end{split}
\end{equation}

\textbf{The sampling differentiable.} Furthermore, to make the sampling process differentiable and ensure the sampling results under a specific distribution, a reparameterization trick is applied \cite{ho2020denoising}: $\bm{x_t}=\bm{\mu_t}+\bm{\sigma_t}\bm{\epsilon_{t-1}}$,  $\bm{\epsilon_{t-1}} \sim \mathcal{N}(\bm{\epsilon_{t-1}}; \bm{0}, \bm{I})$ (this can be verified by the properties of random variable that $\bm{x_t}$ follows a Gaussian distribution with mean $\bm{\mu_t}$ and variance $\sigma_t^2\bm{I}$). Next, we can further simplify to compute   $\bm{x_t}$ by setting $\alpha_t = 1 - \beta_t$, and $\overline{\alpha}_t = \prod_{t=1}^{T}\alpha_t$:

\begin{equation}
    \begin{split}
    \label{supp_f314}
        \bm{x_t}=\sqrt{1-\beta_t}\bm{x_{t-1}} + \sqrt{\beta_t}\bm{\epsilon_{t-1}}\quad\quad\quad\quad\quad\quad\quad\quad\quad\;\;\;\;\;\;\\
=\sqrt{\alpha_t}\bm{x_{t-1}} + \sqrt{1-\alpha_t}\bm{\epsilon_{t-1}}\quad\quad\quad\quad\quad\quad\quad\quad\quad\quad\;\;\;\\
=\sqrt{\alpha_t}(\sqrt{\alpha_{t-1}}\bm{x_{t-2}} + \sqrt{1-\alpha_{t-1}}\bm{\epsilon_{t-1}}) + \sqrt{1-\alpha_t}\bm{\epsilon_t}\;\\ 
=\sqrt{\alpha_t\alpha_{t-1}}\bm{x_{t-2}}+\textcolor{red}{\sqrt{\alpha_t-\alpha_t\alpha_{t-1}}\bm{\epsilon_{t-1}}+\sqrt{1-\alpha_t}\bm{\epsilon_t}}\;\\
\textcolor{blue}{Gaussian \; Variable \;Additivity:} \quad\quad\quad\quad\quad\quad\quad\quad\quad\quad\;\\
\textcolor{blue}{\Rightarrow \epsilon \sim \mathcal{N}(0,(a_t-a_ta_{t-1})+(1-a_t))}\quad\quad\quad\quad\quad\quad\quad\quad\\
=\sqrt{\alpha_t\alpha_{t-1}}\bm{x_{t-2}}+\textcolor{red}{\sqrt{1-\alpha_t\alpha_{t-1}}\bm{\epsilon}}\quad\quad\quad\quad\quad\quad\quad\;\;\;\\
...\quad\quad\quad\quad\quad\quad\quad\quad\quad\quad\quad\quad\quad\quad\quad\quad\quad\quad\quad\quad\quad\;\;\\
=\sqrt{\overline{\alpha}_t}\bm{x_0} + \sqrt{1-\overline{\alpha}_t} \bm{\epsilon}. \quad\quad\quad\quad\quad\quad\quad\quad\quad\quad\quad\;\;\quad\;
    \end{split}
\end{equation}

Therefore, according to Eq.~\ref{supp_f314}, $\bm{x_t}$ is only related to the task target $\bm{x_0}$ and the time label $t$ in the diffusion process, while $\bm{x_t} \sim q(\bm{x_t}|\bm{x_0})=\mathcal{N}(\bm{x_t}; \sqrt{\overline{\alpha}_t}\bm{x_0},({1-\overline{\alpha}_t})\bm{I})$.

That is, during the diffusion process, we can obtain $q(\bm{x_t}|\bm{x_{t-1}})$ and $q(\bm{x_t}|\bm{x_0})$.

\textbf{The inverse of the diffusion process.} The inverse of the diffusion process $q(\bm{x_{t-1}}|\bm{x_t},\bm{x_0})$ does not imply the generation process of DDPMs, which serves as the true Ground Truth in DDPMs. This inverse process is determined by the diffusion process, defining the true posterior distribution that the generation process is required to fit, i.e., $p_\theta(\bm{x_{t-1}}|\bm{x_t},C) \approx q(\bm{x_{t-1}}|\bm{x_t},\bm{x_0})$ (see Fig.~\ref{fig_supp_ddpm}). $q(\bm{x_{t-1}}|\bm{x_t},\bm{x_0})$ also indicates that the computation of the inverse process necessarily involves the task target $\bm{x_0}$. Consistent with the diffusion process, the inverse process also follows i.i.d. and Markov properties. Meanwhile, the inverse process $q(\bm{x_{t-1}}|\bm{x_t},\bm{x_0})$ can directly be calculated by the inverse probability formula (Bayesian formula):

\begin{equation}
\begin{split}
\label{supp_f315}
q(\bm{x_t}|\bm{x_{t-1}},\bm{x_0})=\frac{\textcolor{red}{q(\bm{x_{t-1}}|\bm{x_t},\bm{x_0})}q(\bm{x_t},\bm{x_0})}{q(\bm{x_{t-1}},\bm{x_0})}\quad\quad\quad\\
\textcolor{red}{q(\bm{x_{t-1}}|\bm{x_t},\bm{x_0})}=\frac{q(\bm{x_t}|\bm{x_{t-1}},\bm{x_0})q(\bm{x_{t-1}}|\bm{x_0})}{q(\bm{x_t}|\bm{x_0})}\quad\quad\\
\textcolor{blue}{Markov:} \quad\quad\quad\quad\quad\quad\quad\quad\quad\quad\quad\quad\quad\quad\quad\quad\quad\quad\quad\\
=\frac{\textcolor{black}{q(\bm{x_t}|\bm{x_{t-1}})}q(\bm{x_{t-1}}|\bm{x_0})}{q(\bm{x_t}|\bm{x_0})},\quad\quad\;\;\;\;
\end{split}
\end{equation}
where $q(\bm{x_t}|\bm{x_{t-1}}) = \mathcal{N}(\bm{x_t};\sqrt{\alpha_t}\bm{x_{t-1}},(1-\alpha_t)\bm{I})$, $q(\bm{x_{t-1}}|\bm{x_0}) = \mathcal{N}(\bm{x_{t-1}};\sqrt{\overline{\alpha}_{t-1}}\bm{x_0},(1-\overline{\alpha}_{t-1})\bm{I})$, and $q(\bm{x_t}|\bm{x_0}) = \mathcal{N}(\bm{x_t};\sqrt{\overline{\alpha}_t}\bm{x_0},(1-\overline{\alpha}_t)\bm{I})$. These distributions are known in the derivation of the diffusion process. Subsequently, by substituting $q(\bm{x_t}|\bm{x_{t-1}})$, $q(\bm{x_{t-1}}|\bm{x_0})$ and $q(\bm{x_t}|\bm{x_0})$ into Eq \ref{supp_f315}, the mean $\bm{{\mu_t}}$ and the variance $\sigma_t^2\bm{I}$ of $q(\bm{x_t}|\bm{x_{t-1}},\bm{x_0})=\mathcal{N}(\bm{x_t};\bm{\mu_t}, \sigma_t^2\bm{I})$ can be obtain:

\begin{equation}
\begin{split}
\label{supp_f316}
\bm{\mu_t}=\frac{\sqrt{\alpha_t}(1-\overline{\alpha}_{t-1})}{1-\overline{\alpha_t}}\bm{x_t}+\frac{\sqrt{\overline{\alpha}_{t-1}}(1-\alpha_t)}{1-\overline{\alpha}_t}\bm{x_0},\\
\bm{\sigma_t^2}=\frac{1-\overline{\alpha}_{t-1}}{1-\overline{\alpha}_t}(1-\alpha_t)\bm{I}.\quad\quad\quad\quad\quad\quad\quad\quad\;
\end{split}
\end{equation}

Meanwhile, in Eq.~\ref{supp_f316}, we can observe that the variance $\sigma_t^2\bm{I}$ is a constant term (some works also set the variance as a fitting term \cite{nichol2021improved, fan2023optimizing}). 

Next, due to the better performance observed in experiment \cite{ho2020denoising}, $\bm{x_0}$ is considered to be replaced by $\bm{\epsilon}$, i.e., $\bm{x_0}=\frac{\bm{x_t}-\sqrt{1-\overline{\alpha}_t}\bm{\epsilon}}{\sqrt{\overline{\alpha}_t}}$:

\begin{equation}
\begin{split}
\label{supp_f317}
\bm{\mu_t}=\frac{1}{\sqrt{\alpha_t}}(\bm{x_t}-\frac{1-\alpha_t}{\sqrt{1-\overline{\alpha}_t}}\bm{\epsilon}).
\end{split}
\end{equation}

Notably, the initial value $\bm{x_T}$ of $\bm{x_t}$ can be directly sampled from a prior distribution $P_{noise}$ (Gaussian distribution, $\bm{x_T} 
\sim \mathcal{N}(\bm{x_T};\bm{0},\bm{I})$) during inference. Therefore, the only unknown term is $\bm{\epsilon}$ in Eq.~\ref{supp_f317}. 



\subsection{Generation Process} 
\label{sec3.2}

\textbf{The fitting target.} The generation process defines the generation mode of DDPMs: unconditional generation and  conditional generation. This takes the inverse of the diffusion process as the fitting target, i.e., $p_\theta(\bm{x_{t-1}}|\bm{x_t},C) \approx q(\bm{x_{t-1}}|\bm{x_t},\bm{x_0})$. To better fit the inverse process, each step of the generation process is characterized by i.i.d. and Markov properties. 

We directly derive the fitting target of conditional DDPMs, as unconditional generation ($\bm{c}=\emptyset$) can be viewed as a special case of conditional DDPMs solely conditioned on the time label $t$ ($C=\{\bm{c},t\}$).

\textbf{The detailed derivation.} Formally, given a set of conditions $C=\{\bm{c_i},t|i=1..n\}$ ("$n$" means the number of conditions, this means that conditional DDPMs can perform the multi-conditional generation), we can compute the reverse process by the joint distribution $p_\theta(\bm{x_{0:T}},C)$:

\begin{equation*}
\begin{split}
p_\theta(\bm{x_{0:T}},C)=\textcolor{red}{p_\theta(\bm{x_0}|\bm{x_{1:T}},C)}p_\theta(\bm{x_{1:T}},C)\quad\quad\quad\quad\quad\quad\;\\
\textcolor{blue}{Markov:} \quad\quad\quad\quad\quad\quad\quad\quad\quad\quad\quad\quad\quad\quad\quad\quad\quad\quad\quad\quad\\
=\textcolor{red}{p_\theta(\bm{x_0}|\bm{x_1},C)}p_\theta(\bm{x_1}|\bm{x_{2:T}},C)p_\theta(\bm{x_{2:T}},C)\;\;\\ 
\quad\quad=p_\theta(\bm{x_0}|\bm{x_1},C)...p_\theta(\bm{x_T}|\bm{x_{T-1}},C)p(\bm{x_T},C)\;\\ 
=p(\bm{x_T},C)\prod_{t=1}^{T}p_\theta(\bm{x_{t-1}}|\bm{x_t},C)\quad\quad\quad\quad\quad\;\;\\ 
\end{split}
\end{equation*}

\begin{equation}
\begin{split}
	\label{supp_f321}
\textcolor{blue}{\bm{x_T} \sim \mathcal{N}(\bm{x_T};0,I)}\quad\quad\quad\quad\quad\quad\quad\quad\quad\quad\quad\quad\quad\quad\quad\quad\\ 
=p(\bm{x_T})\prod_{t=1}^{T}p_\theta(\bm{x_{t-1}}|\bm{x_t},C),\quad\quad\quad\quad\quad\quad\;\;\\
\end{split}
\end{equation}
where $p_\theta(\bm{x_{t-1}}|\bm{x_t},C)=\mathcal{N}(\bm{x_{t-1}};{\mu_\theta(\bm{x_t},C)},\sigma_t^2\bm{I})$, as in Eq.~\ref{supp_f317}, the mean $\bm{\mu_t}$  contains unknown variables during inference, while the variance factor $\sigma_t^2$ is a constant term. 


Next, according to Eq.~\ref{supp_f317}, we can logically further refine this fitting target:

\begin{equation}
\begin{split}
	\label{supp_f322}
    p_\theta(\bm{x_{t-1}}|\bm{x_t},C) \approx q(\bm{x_{t-1}}|\bm{x_t},\bm{x_0}), \\
    \Rightarrow {\mu_\theta(\bm{x_t},C)} \approx \bm{\mu_t}, \quad \quad  \quad  \quad \quad \quad \; \;\; \; \; \; \\
    \Rightarrow \epsilon_\theta(\bm{x_t},C) \approx \bm{\epsilon}.\quad \quad  \quad  \quad \quad \quad\;\;\;\; \;\;\;\;\;
\end{split}
\end{equation}

According to Eq.~\ref{supp_f322}, we can clearly recognize that as long as the generation process can sufficiently fit the inverse of the diffusion process, DDPMs can achieve the multi-condition generation. Meanwhile, due to the inverse process with a total of $T$ steps, the final training objective is:


\begin{equation}
\begin{split}
\label{supp_f323}
L(\theta)=\frac{1}{T}\sum_{t=1}^{T} D_{KL}(q(\bm{x_{t-1}}|\bm{x_t},\bm{x_0})||p_\theta(\bm{x_{t-1}}|\bm{x_t},C)),\\
=\frac{1}{T}\sum_{t=1}^{T}||\bm{\mu_t} - \mu_\theta(\bm{x_t},C)||^2,\quad\quad\quad\quad\quad\quad\quad\;\;\;\;\;\\
 =\mathbb{E}_{\bm{\epsilon} \sim \mathcal{N}(\bm{0},\bm{I})}||\bm{\epsilon} - \epsilon_\theta(\bm{x_t},C)||^2.\quad\quad\quad\quad\quad\quad\;\;\;\;\;\;
\end{split}
\end{equation}

Thus, we derive Eq.~1 of the main text. 

\textbf{The intuitive explanation.} Similarly, we can also understand the generation process in a more intuitive way.
The generation process aims to fit the reverse process to achieve the  generalization of generation, due to the computation of $q(\bm{x_{t-1}}|\bm{x_t},\bm{x_0})$ involving $\bm{x_0}$. According to Eq.~\ref{supp_f315} 
and Eq.~\ref{supp_f317}, the fitting target of the generation process can conduct two stages of simplification: fitting the distribution $q(\bm{x_{t-1}}|\bm{x_t},\bm{x_0})$ $\rightarrow$ fitting the distribution hyperparameter $\bm{\mu_t}$ $\rightarrow$ fitting the unknown variable $\bm{\epsilon}$. Meanwhile, due to the inverse process with a total of $T$ steps, thus the fitting objective:

\vspace{-8pt}
\begin{equation}
\begin{split}
\label{supp_f324}
L(\theta)=\frac{1}{T}\sum_{t=1}^{T}||\bm{\epsilon} - \epsilon_\theta(\bm{x_t},C)||^2. 
\end{split}
\end{equation}

For a more intuitive description, we express the expectation as a summation in Eq.~\ref{supp_f323}.

\subsection{How do we introduce DDPMs to 3D tasks?}
\label{sec3.3}

\begin{figure}[htp]
	\centering
	\includegraphics[width=0.48\textwidth]
 {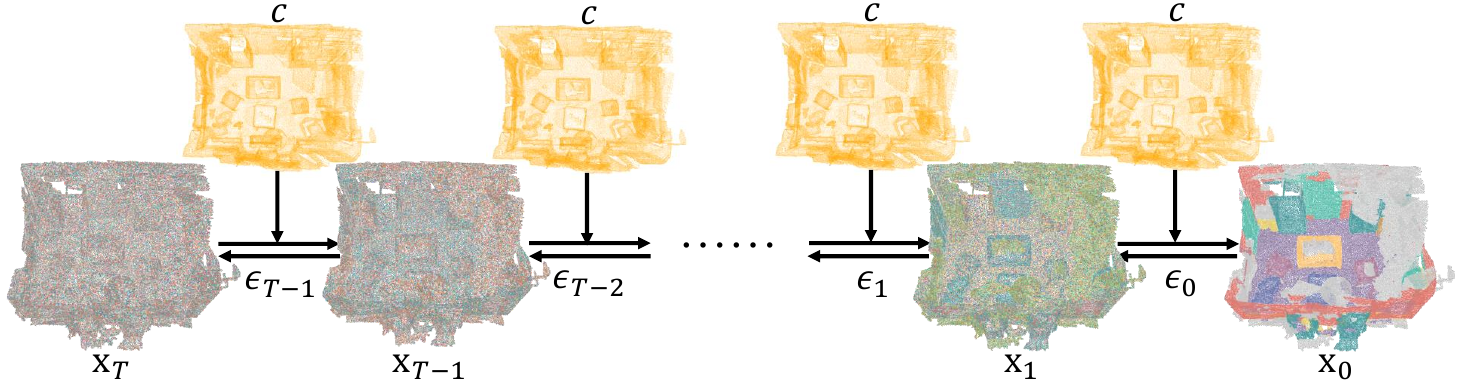}
 
	\caption{Applying DDPMs to the point cloud semantic segmentation task: the semantic label $\bm{x_0}$ is gradually perturbed with noise during the diffusion process and slowly reconstructed during the generation process conditioned on the segmented point cloud $\bm{c}$.}
	\label{fig_supp_pcss}
\end{figure}

According to Sec.~\ref{sec3.1} and Sec.~\ref{sec3.2}, we can apply DDPMs to most existing 3D tasks. For example, in the point cloud semantic segmentation task, $\bm{c}$ means the segmented point cloud, and $\bm{x_0}$ represents the semantic label (see Fig.~\ref{fig_supp_pcss}). In the diffusion process, the semantic label $\bm{x_0}$ is gradually noised until $\bm{x_0}$ degenerates into $\bm{z}$. Meanwhile, in the generation process, the semantic label $\bm{x_0}$ is gradually reconstructed until $\bm{z}$ is restored to the desired $\bm{x_0}$ under the condition of the segmented point cloud $\bm{c}$. This achieves point cloud semantic segmentation tasks. DDPMs also can be introduced into other 3D tasks in a similar manner.

\subsection{Why is the performance of DDPMs determined by the noise fitting quality?} 
\label{sec3.4}

According to Eq.~\ref{supp_f315}, the generation process of DDPMs fits the reverse of the diffusion process at all time steps. Meanwhile, in Eq.~\ref{supp_f317},  the noise $\bm{\epsilon}$ is an unknown in distribution mean $\bm{\mu_t}$. Therefore, the generation process of DDPMs essentially fits the unknown noise $\bm{\epsilon}$ (see Eq.~\ref{supp_f322}). This means that the higher the the noise fitting quality, the better the generation result of DDPMs will be.

\begin{figure*}[htp]
	\centering
	\includegraphics[width=\textwidth]
 {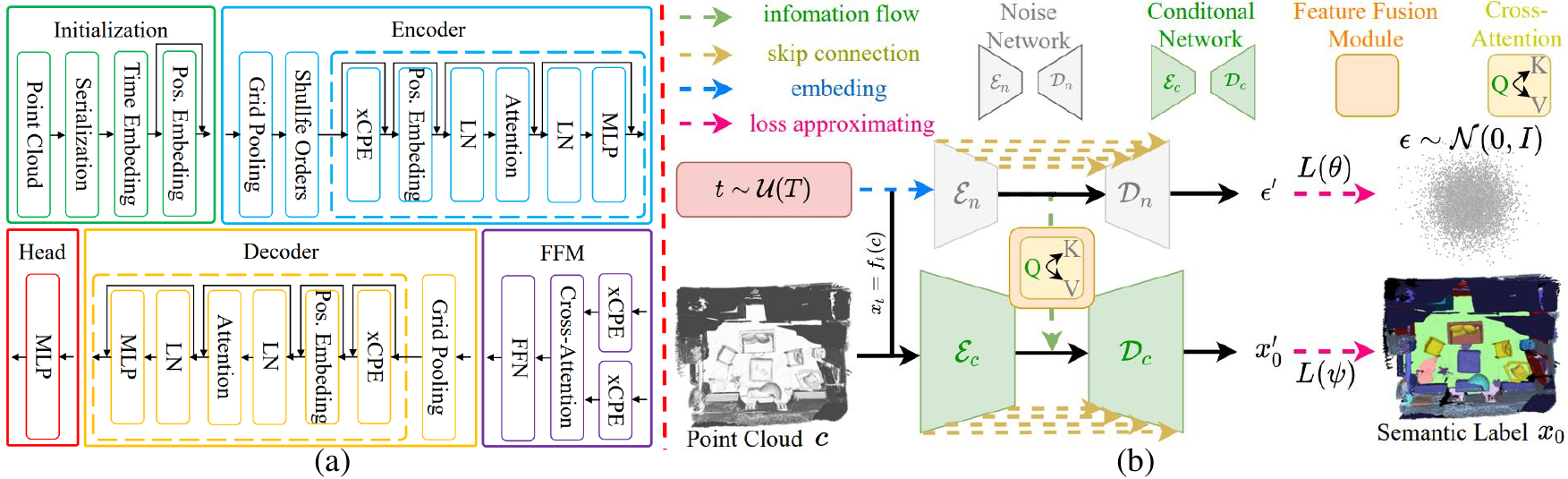}
 \vspace{-0.6cm} 
	\caption{(a) describes the five components that make up CDSegNet: the Initialization Module, the Encoder, the Feature Fusion Module (FFM), the Decoder, and the Head. Meanwhile, (b) shows the overall framework of CDSegNet, where both the Noise Network (NN) and the Condition Network (CN) consist of three parts: the Encoder, the Decoder, and the Head.}
	\label{fig_supp_network}
\end{figure*}

\subsection{Why are DDPMs robust to noise in the modeled distribution?} 
\label{sec3.5}

\textbf{From the gradient of the data distribution.} \cite{qu2024conditional} provides an intuitive explanation from the gradient of data distribution. Under stochastic differential equations (SDEs), the target noise in conditional DDPMs can be converted into and from the score that means the gradient of data distribution  by a constant factor $\alpha = -\frac{1}{\sqrt{1-\overline{\alpha}_t}}$ \cite{song2021scorebased}:

\vspace{-5pt}
\begin{equation}
\begin{split}
	\label{supp_f341}
	\alpha \epsilon_\theta(\bm{x_t},C) = s_\theta\bm{(x_t},C) \approx \nabla_{\bm{x_t}} \log P_t(\bm{x_t}).
\end{split}
\end{equation}

This means that the predicted noise guides the transformation between the two distributions (see Fig.~\ref{fig_supp_ddpm}), i.e., $\bm{x_t} \stackrel{\bm{\epsilon_{t-1}}}{\longrightarrow} \bm{x_{t-1}}$, $\bm{\epsilon_{t-1}}=\epsilon_\theta(\bm{x_t},C) \sim P_{noise}(\bm{\epsilon_{t-1}}|\bm{x_t},C)$. Perturbing $C$ inevitably affects the generation quality of $\bm{\epsilon_{t-1}}$, thus directly affecting the quality of the transformation between $\bm{x_t}$ and $\bm{x_{t-1}}$. When the distribution of the perturbation is identical or similar to the distribution of $\bm{\epsilon_{t-1}}$, the impact on the generation of $\bm{\epsilon_{t-1}}$ is relatively small; otherwise, it is greater. Therefore, DDPMs are robust to the modeled distribution noise. Meanwhile, the closer the noise distribution is to the modeled distribution, the better the noise robustness; otherwise, this deteriorates.



\textbf{From the noise samples and the noise fitting.} In the main text, we provide a simpler explanation that is consistent with the source. Since DDPMs can see multi-level noise samples $\bm{x_t}$  and fit the noise target $\bm{\epsilon}$ from the modeled distribution during training, DDPMs can adapt to related distribution noise during inference compared to non-DDPMs.

\subsection{Why DDPMs require more training and inference iterations than non-DDPMs?}
\label{sec3.6}

This is because DDPMs require fitting more intermediate samples than non-DDPMs. For a sample pair $\{\bm{c},\bm{x_0}\}$, the training object of DDPMs is:

\vspace{-8pt}
\begin{equation}
\begin{split}
	\label{supp_f361}
	L_\theta =\frac{1}{T}\sum_{t=1}^{T}||\bm{y_{t-1}}-f_\theta(\bm{x_{t}}, \bm{c})||^2,
\end{split}
\end{equation}
where $f_\theta$ indicates a neural network with sufficient fitting ability.  $\bm{y_{t-1}}$ represents $\bm{\epsilon}$ at corresponding the time label $t$ in Eq.~\ref{supp_f324} (we omit the time label $t$ as part of the input). This means that DDPMs require fitting $T$ targets, i.e. $\bm{x_0}=\{\bm{y_{t-1}}|t=1...T\}$, due to the significant error of fitting distributions with a large difference when using a single step or a small number of steps \cite{song2021scorebased, song2020denoising, lu2022dpm}.

Meanwhile, this also makes that DDPMs require to iterate $T$ steps to achieve the accurate result during inference:
\vspace{-20pt}

\begin{equation}
\begin{split}
	\label{supp_f362}
	 \bm{y_{t-1}^{'}}=f_\theta(\bm{x_{t}},\bm{c}), \quad t \in [1,T], 
\end{split}
\end{equation}
where $\bm{y_{t-1}^{'}}$ means the predicted noise in DDPMs. According to Eq.~\ref{supp_f362}, DDPMs require $T$ steps to converge ($\{\bm{y'_{i-1}}|i=1...T\}$).

However, non-DDPMs only necessitate one step (the input $\bm{x_{t}}$=$\emptyset$, while $T$=1) for the training and inference:

\vspace{-8pt}
\begin{equation}
\begin{split}
	\label{supp_f363}
	L_{\theta} =||\bm{y_{0}}-f_\theta(\emptyset, \bm{c})||^2, \quad \bm{y_{0}^{'}}=f_\theta(\emptyset,\bm{c}),
\end{split}
\end{equation}
where the target $\bm{y_{0}}$=$\bm{x_0}$, while $\bm{y_{0}^{'}}$ means the predicted $\bm{x_0}$.

Therefore, under the same setting, DDPMs are bound to conduct more training and inference iterations than non-DDPMs.

\subsection{Why can the CNF of DDPMs achieve single-step inference?}
\label{sec3.7}

According to Sec.~\ref{sec3.4}, the performance of DDPMs depends on the noise fitting quality in NN. This necessitates multiple iterations, as the significant errors occur in a single step. Although reducing the number of iterations can accelerate the sampling process \cite{song2020denoising, lu2022dpm}, this introduces two limitations: a loss of accuracy \cite{song2021scorebased, song2020denoising, lu2022dpm} and a decrease in diversity (less noise introduced resulting in less randomness \cite{song2020denoising}). Fortunately, except generation tasks, most 3D tasks focus on the certainty of the results. CNF considers CN rather than NN as the task-dominant network, cleverly avoiding extensive iterations from DDPMs. Meanwhile, since NN still see noise samples and fits the noise during training, models retain the robustness from DDPMs.

We do not deny NCF of DDPMs, which thrives in generative tasks. Our goal is to provide a new way of applying DDPMs to 3D tasks, hoping to expand the application scope of DDPMs. In fact, generation tasks require diversity, while other tasks focus on certainty. CNF actually sacrifices the result diversity to enhance the performance certainty.


\begin{figure*}[htp]
	\centering
	\includegraphics[width=\textwidth]
 {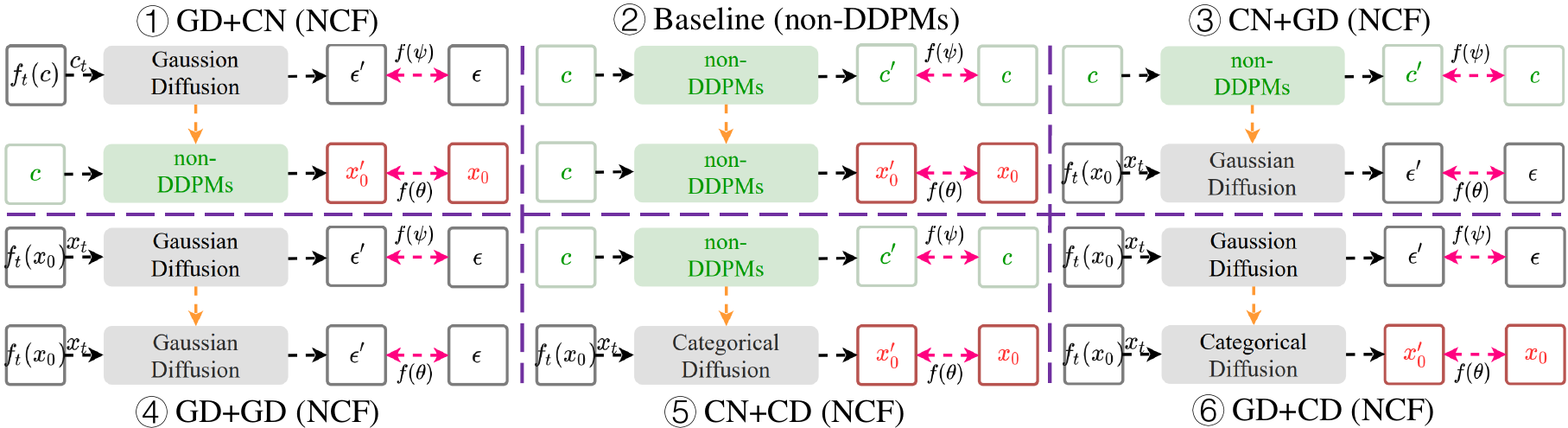}
 \vspace{-0.6cm} 
	\caption{\textcircled{1} GD+CD (the auxiliary network as a Gaussian diffusion, the dominant network as a non-DDPM, CNF), \textcircled{2} Baseline (the auxiliary network as a non-DDPM, the dominant network as a non-DDPM, non-DDPMs), \textcircled{3} CN+GD (the auxiliary network as a non-DDPM, the dominant network as a Gaussian diffusion, NCF), \textcircled{4} GD+GD (the auxiliary network as a Gaussian diffusion, the dominant network as a Gaussian diffusion, NCF), \textcircled{5} CN+CD (the auxiliary network as a non-DDPM, the dominant network as a categorical diffusion, NCF), and \textcircled{6} GD+CD (the auxiliary network as a Gaussian diffusion, the dominant network as a categorical diffusion, NCF).}
	\label{fig_supp_combination}
\end{figure*}

\section{Implementation}
\label{sec4}

\subsection{Model Hyperparameters}
\label{sec4.1}

In this section, we describe the implementation details of CDSegNet. CDSegNet is built on top of PTv3 and is depicted in Fig.~\ref{fig_supp_network}. For subsequent pooling and denoising, the initialization module serializes the point cloud and conducts the time and position embeding. Meanwhile, both the Noise Network (NN) and the Conditional Network (CN), each composed of an encoder, a decoder, and a head, fit the noise and the task target (semantic labels), respectively. Moreover, FFM directs the noise information from NN to CN, enhancing the semantic features in CN. The detailed parameters of the network architecture of CDSegNet are described in Tab.~\ref{supp_tab411}, while the training hyperparameters for each benchmark (ScanNet, ScanNet200, nuScenes) are shown in Tab.~\ref{supp_tab412}.

We used 4 NVIDIA 4090 GPUs to train CDSegNet on ScanNet, ScanNet200 and nuScenes, which took approximately 21 hours, 21 hours and 29 hours, respectively. Additionally, we also tried to train CDSegNet on ScanNet using a single NVIDIA 3090 GPU, which took approximately 65 hours (batch size=2, this still can achieve 77.9 mIoU).

\begin{table}[h]
        \scriptsize
  \resizebox{0.48\textwidth}{!}{
	\begin{tabular}{p{3.0cm}p{3.0cm}}	
       \Xhline{1pt}
  
        Config
        &\makecell[c]{Parameter} \\
        \cline{1-2}
        
        Serialization Pattern
        &\makecell[c]{Z + TZ + H + TH}\\

        Patch Interaction
        &\makecell[c]{Shift Order + Shuffle Order}\\

        Time Embeding
        &\makecell[c]{Cos-Sin (128)}\\

        Positional Embeding
        &\makecell[c]{xCPE (32)}\\

        MLP Ratio
        &\makecell[c]{4}\\

        QKV Bias
        &\makecell[c]{True}\\

        Drop Path
        &\makecell[c]{0.3}\\

        \hline

        NN Stride
        &\makecell[c]{[4,4]}\\

        NN Encoder Depth
        &\makecell[c]{[2,2]}\\

        NN Encoder Channels
        &\makecell[c]{[64,128]}\\

        NN Encoder Num Heads
        &\makecell[c]{[4,8]}\\

        NN Encoder Patch Size
        &\makecell[c]{[1024,1024]}\\

        NN Decoder Depth
        &\makecell[c]{[2,2]}\\

        NN Decoder Channels
        &\makecell[c]{[64,64]}\\

        NN Decoder Num Heads
        &\makecell[c]{[4,4]}\\

        NN Decoder Patch Size
        &\makecell[c]{[1024,1024]}\\

        NN Skip Connection
        &\makecell[c]{Element Addition}\\
        
        \hline

        CN Stride
        &\makecell[c]{[2,2,2,2]}\\

        CN Encoder Depth
        &\makecell[c]{[2,2,6,6]}\\

        CN Encoder Channels
        &\makecell[c]{[64,128,256,512]}\\

        CN Encoder Num Heads
        &\makecell[c]{[4,8,16,32]}\\

        CN Encoder Patch Size
        &\makecell[c]{[1024,1024,1024,1024]}\\

        CN Decoder Depth
        &\makecell[c]{[2,2,2,2]}\\

        CN Decoder Channels
        &\makecell[c]{[64,64,128,256]}\\

        CN Decoder Num Heads
        &\makecell[c]{[4,4,8,16]}\\

        CN Decoder Patch Size
        &\makecell[c]{[1024,1024,1024,1024]}\\

        CN Skip Connection
        &\makecell[c]{Channel Concatenation}\\

        \hline

        FFM Position Encoding
        &\makecell[c]{xCPE (128,512)}\\

        FFM Feat Scale
        &\makecell[c]{1.0}\\

        FFM Depth
        &\makecell[c]{[1,]}\\

        FFM Channels
        &\makecell[c]{[512,]}\\

        FFM Num Heads
        &\makecell[c]{[32,]}\\

        FFM Patch Size
        &\makecell[c]{[1024,]}\\
        
        \Xhline{1pt}
        
	\end{tabular}
}
	\caption{The parameters of network framework for CDSegNet.}
	\label{supp_tab411}

\end{table}

\begin{table}[h]
        \scriptsize
  \resizebox{0.48\textwidth}{!}{
	\begin{tabular}{p{1.1cm}p{1.2cm}p{0.0005cm}p{1.1cm}p{1.2cm}p{0.0005cm}p{1.1cm}p{1.2cm}}	
       \Xhline{1pt}

        \multicolumn{2}{c}{{ScanNet} \cite{dai2017scannet}}
        &\quad
        &\multicolumn{2}{c}{{ScanNet200} \cite{rozenberszki2022language}}
        &\quad
        &\multicolumn{2}{c}{{nuScenes} \cite{caesar2020nuscenes}}
        \\
        \cline{1-2} \cline{4-5} \cline{7-8} 
        
        {Config}
        &\makecell[c]{{Parameter}} 
        &\quad
        &{Config}
        &\makecell[c]{{Parameter}}
        &\quad
        &{Config}
        &\makecell[c]{{Parameter}}
        \\
        \hline

        Optimizer
        &\makecell[c]{AdamW} 
        &\quad
        &Optimizer
        &\makecell[c]{AdamW}
        &\quad
        &Optimizer
        &\makecell[c]{AdamW}
        \\

        Scheduler
        &\makecell[c]{Cosine} 
        &\quad
        &Scheduler
        &\makecell[c]{Cosine}
        &\quad
        &Scheduler
        &\makecell[c]{Cosine}
        \\

        LR
        &\makecell[c]{0.002} 
        &\quad
        &LR
        &\makecell[c]{0.002}
        &\quad
        &LR
        &\makecell[c]{0.002}
        \\

        Block LR
        &\makecell[c]{0.0002} 
        &\quad
        &Block LR
        &\makecell[c]{0.0002}
        &\quad
        &Block LR
        &\makecell[c]{0.0002}
        \\

        Weight De.
        &\makecell[c]{0.005} 
        &\quad
        &Weight De.
        &\makecell[c]{0.005}
        &\quad
        &Weight De.
        &\makecell[c]{0.005}
        \\

        Batch Size
        &\makecell[c]{8} 
        &\quad
        &Batch Size
        &\makecell[c]{8}
        &\quad
        &Batch Size
        &\makecell[c]{8}
        \\

        Epoch
        &\makecell[c]{800} 
        &\quad
        &Epoch
        &\makecell[c]{800}
        &\quad
        &Epoch
        &\makecell[c]{50}
        \\
        \hline
        
        Loss St.
        &\makecell[c]{GLS} 
        &\quad
        &Loss St.
        &\makecell[c]{GLS}
        &\quad
        &Loss St.
        &\makecell[c]{GLS}\\

        Task Num
        &\makecell[c]{2} 
        &\quad
        &Task Num
        &\makecell[c]{2}
        &\quad
        &Task Num
        &\makecell[c]{2}
        \\
        \hline

        Target
        &\makecell[c]{$\epsilon$} 
        &\quad
        &Target
        &\makecell[c]{$\epsilon$}
        &\quad
        &Target
        &\makecell[c]{$\epsilon$}
        \\

        T
        &\makecell[c]{1000} 
        &\quad
        &T
        &\makecell[c]{1000}
        &\quad
        &T
        &\makecell[c]{1000}
        \\

        Schedule
        &\makecell[c]{Cosine} 
        &\quad
        &Schedule
        &\makecell[c]{Linear}
        &\quad
        &Schedule
        &\makecell[c]{Linear}
        \\

        Range
        &\makecell[c]{[0,1000]} 
        &\quad
        &Range
        &\makecell[c]{[1e-1,1e-5]}
        &\quad
        &Range
        &\makecell[c]{[1e-2,1e-3]}
        \\
        
        \Xhline{1pt}
        
	\end{tabular}
}
	\caption{The training hyperparameters of CDSegNet for different benchmarks.}
	\label{supp_tab412}

\end{table}

\begin{figure}[htp]
	\centering	\includegraphics[width=0.48\textwidth]
 {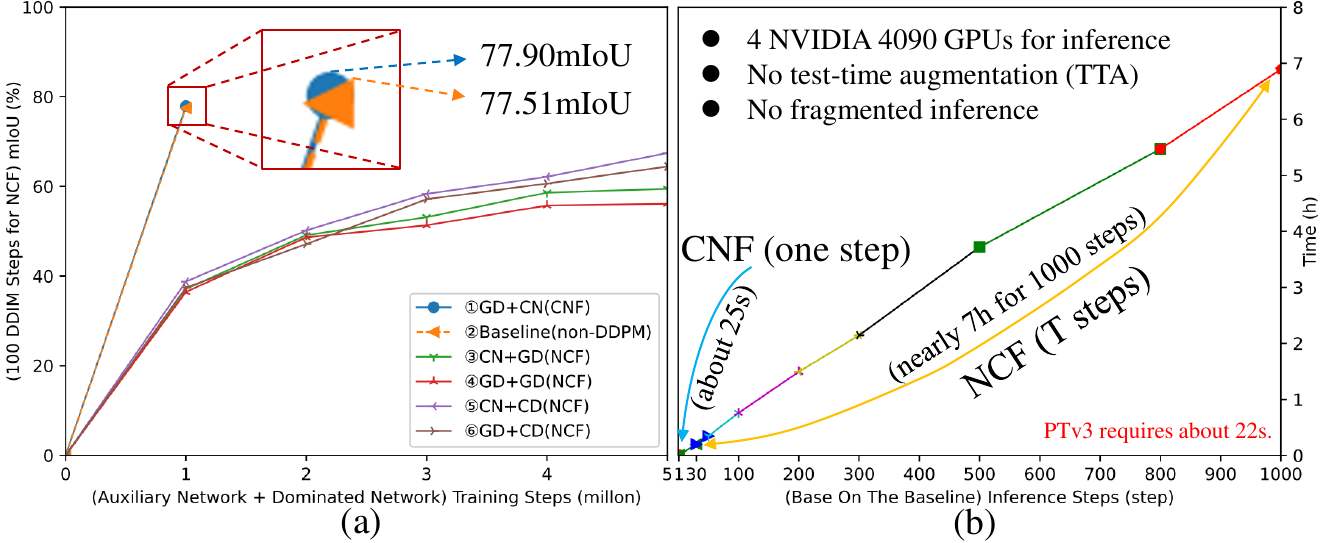}
	\caption{We try several combinations for conditional DDPMs built on the baseline  on ScanNet in (a). The model of CNF (\small \textcircled{1}) and baseline (\small \textcircled{2}) only perform a step inference. Meanwhile, the  models of NCF (\small \textcircled{3},\small \textcircled{4},\small \textcircled{5},\small \textcircled{6}) use DDIM for 100 inference steps. (b) shows the inference time cost of CNF and NCF under the baseline. To effectively evaluate the inference cost, the models of CNF and NCF do not use test augmentation, and the voxel size of the input point cloud is 0.001m (i.e., no fragmentation inference). CNF achieves better performance with fewer iterations.}
	\label{fig_supp_training_inference}
\end{figure}

\subsection{Combinations for Conditional DDPMs}

As mentioned in Sec.~4.1 of the main text, CN and NN allow us to transcend the limitation of non-DDPMs and DDPMs. This means that NCF and CNF can use any type of DDPMs \cite{bansal2024cold}. This only requires CN (the non-DDPM process) to be the dominant backbone in CNF, while NCF is dominated by NN (the DDPM process) (see Fig.~\ref{fig_supp_combination} and Fig.~\ref{fig_supp_training_inference}). For example, in Fig.~\ref{fig_supp_training_inference} of the main text, CD+DD (NCF) indicates that NN and CN are modeled as a Gaussian diffusion \cite{ho2020denoising} and a categorical diffusion \cite{austin2021structured}, respectively. Therefore, this can produce various combinations of conditional DDPMs. The above combination (NCF, \textcircled{3}, \textcircled{4}, \textcircled{5}, \textcircled{6}) is the same as the network framework of Baseline (\textcircled{2}) and CDSegNet (CNF, \textcircled{1}). We provide   \href{https://github.com/QWTforGithub/CDSegNet/blob/main/pointcept/models/default.py}{implementations} for all combinations.

For modeling the Gaussian distribution, we follow \cite{ho2020denoising}, which simply employs MSE loss in NN to approximate the noise $\bm{\epsilon}$. Meanwhile, for modeling the categorical distribution, $\bm{x_0}$ is used as the fitting target, and we employ KL divergence loss and cross-entropy loss in NN, similar to \cite{austin2021structured}. 

\subsection{Integration of CNF into PointNet and PointNet++}

\begin{figure}[htp]
	\centering
	\includegraphics[width=0.48\textwidth]
 {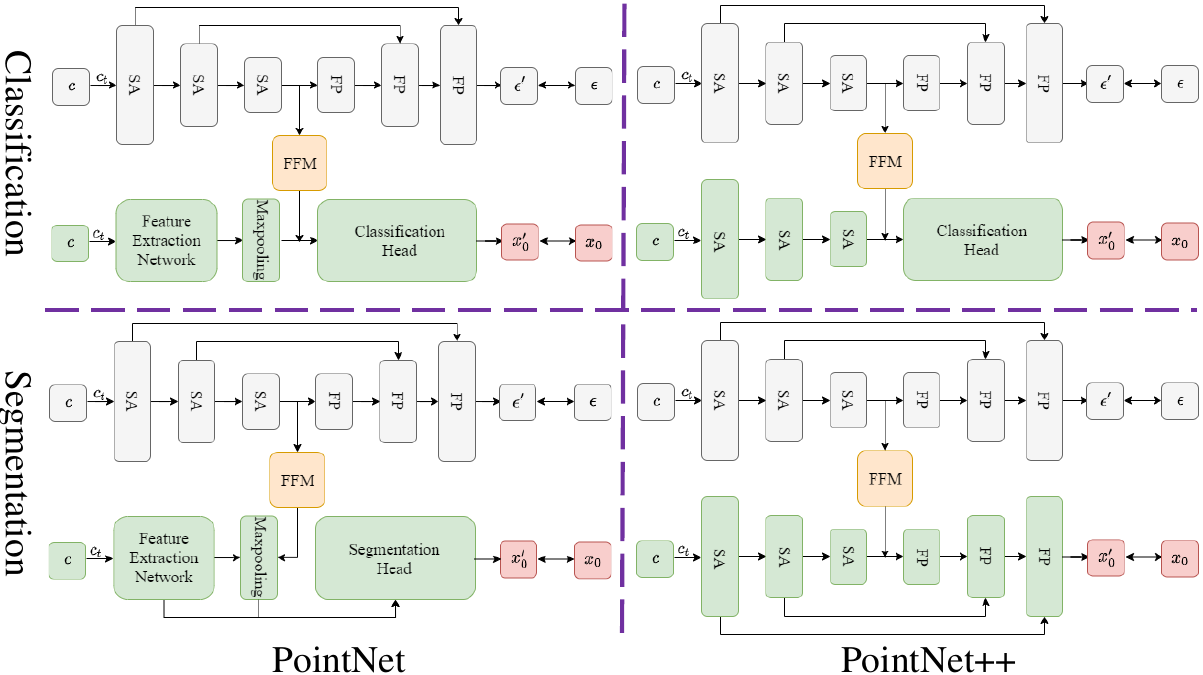}
 \vspace{-0.5cm} 
	\caption{The framework of applying CNF to PointNet and PointNet++. We use a additional PointNet++ to model the diffusion process and FFM of CDSegNet as a noise filter.}
	\label{fig_supp_pointnet_pointnet++}
\end{figure}

Fig.~\ref{fig_supp_pointnet_pointnet++} illustrates the framework of introducing CNF to PointNet and PointNet++. We use a additional PointNet++ to model the diffusion process and employ the FFM of CDSegNet as a noise filter. For PointNet, CNF is applied after the feature pooling stage. Meanwhile, for PointNet++, we introduce CNF at the bottleneck stage of the U-Net.

\section{Optimization for CNF}
\label{sec5}

Benefiting from a dual-branch framework, CNF has two fitting objectives: the noise fitting (NN) and the task-target fitting (CN). Therefore, CNF can be optimized from two perspectives: DDPMs and multi-task learning. Using PTv3+CNF on nuScenes as an example, we gradually exhibit the entire optimization process of CNF, aiming to provide a guidance for future applications (\textbf{All optimization processes are from previous checkpoint. However, this does not affect the performance improvement results}). 

\textbf{Baseline.} Our baseline is PTv3+CNF, which is trained on nuScenes. This simply uses PTv3 as  CN, with the additional inclusion of FFM and NN of CDSegNet. The network architecture and hyperparameter settings of CN in PTv3+CNF are entirely consistent with PTv3. 

Tab.~\ref{supp_tab51} shows the comparison results between the baseline and PTv3. The baseline (this directly introduces CNF onto PTv3 without any optimization) demonstrates a significant improvement in noise robustness. This is because the multi-level feature perturbations from NN enhance the noise adaptability of PTv3. Meanwhile, the baseline performs worse  than PTv3 in terms of overall performance. This means that unreasonable noise perturbations harm the performance of CN.

\begin{table}[h]
        \scriptsize
  \resizebox{0.48\textwidth}{!}{
	\begin{tabular}{p{1.0cm}p{0.8cm}p{0.8cm}p{0.8cm}p{0.005cm}p{1.2cm}}	
        \Xhline{1pt}
  
        \multirow{2}{*}{Methods}
        &\multicolumn{3}{c}{Performance} 
        &\quad
        &\makecell[c]{Robustness}  \\
        \cline{2-4} \cline{6-6}
        
        &\makecell[c]{mIoU}
        &\makecell[c]{mAcc}
        &\makecell[c]{allAcc}
        &\quad
        &\makecell[c]{$\tau$=0.1}\\
       \hline

        PTv3 \cite{wu2024point}
        &\makecell[c]{\textbf{80.3}}
        &\makecell[c]{\textbf{87.2}}
        &\makecell[c]{\textbf{94.6}}
        &\quad
        &\makecell[c]{{63.9}}\\

        \rowcolor{gray!20}
        Baseline
        &\makecell[c]{{79.6}}
        &\makecell[c]{{86.9}}
        &\makecell[c]{{94.1}}
        &\quad
        &\makecell[c]{\textbf{66.8}}\\

        \Xhline{1pt}
        
	\end{tabular}
 }
	\caption{The results of the baseline and PTv3 on nuScenes.}
	\label{supp_tab51}
\end{table}

\textbf{The skip connection mode in the Decoder.} We experimented with different skip connection modes in the Decoder: Element-Wise Addition (Baseline), Element-Wise Multiplication (EWM), and Channel Concatenation (CC). 

Tab.~\ref{supp_tab52} shows that CC achieves the better performance. Some works \cite{huang2022scale, si2023freeu} have demonstrated that in DDPMs, the Decoder typically generates high-frequency information, i.e., the details of the generated results. This requires feature fusion to retain as much information as possible. Channel concatenation effectively preserves information from the skip features and the backbone features though the expansion of the channel dimension. However, element-wise addition and multiplication causes the elements of the skip features and the backbone features at the same spatial location to share the same number of channels, which may limit the ability of models to capture details.

We chose the model with the skip connection mode CC as the baseline.

\begin{table}[h]
        \scriptsize
  \resizebox{0.48\textwidth}{!}{
	\begin{tabular}{p{1.0cm}p{0.8cm}p{0.8cm}p{0.8cm}p{0.005cm}p{1.2cm}}	
        \Xhline{1pt}
  
        \multirow{2}{*}{Methods}
        &\multicolumn{3}{c}{Performance} 
        &\quad
        &\makecell[c]{Robustness}  \\
        \cline{2-4} \cline{6-6}
        
        &\makecell[c]{mIoU}
        &\makecell[c]{mAcc}
        &\makecell[c]{allAcc}
        &\quad
        &\makecell[c]{$\tau$=0.1}\\
       \hline

        Baseline
        &\makecell[c]{\underline{79.6}}
        &\makecell[c]{\underline{86.9}}
        &\makecell[c]{\underline{94.1}}
        &\quad
        &\makecell[c]{\underline{66.9}}\\

        EWM
        &\makecell[c]{{79.3}}
        &\makecell[c]{{85.9}}
        &\makecell[c]{{93.5}}
        &\quad
        &\makecell[c]{{66.5}}\\

        \rowcolor{gray!20}
        CC
        &\makecell[c]{\textbf{79.8}}
        &\makecell[c]{\textbf{87.0}}
        &\makecell[c]{\textbf{94.2}}
        &\quad
        &\makecell[c]{\textbf{67.1}}\\

        \Xhline{1pt}
        
	\end{tabular}
 }
	\caption{The results of the different skip connection modes in the Decoder on nuScenes.}
	\label{supp_tab52}

\end{table}

\begin{figure}[htp]
	\centering
	\includegraphics[width=0.48\textwidth]{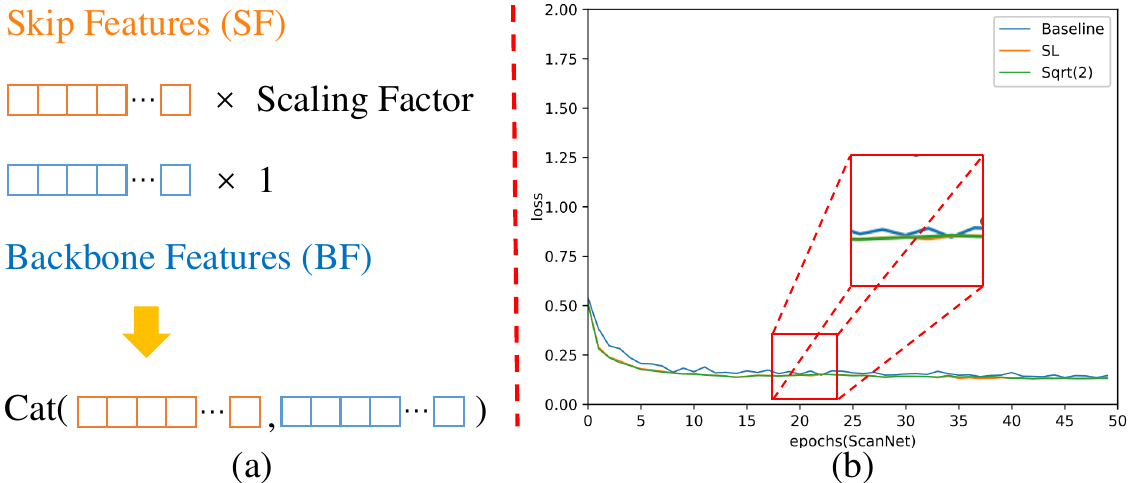}
 \vspace{-0.5cm} 
	\caption{(a) shows the detail of the skip feature scaling. Meanwhile, (b) demonstrates a comparison of the training loss between the baseline, the skip feature scaling SL (ScaleLong) \cite{huang2022scale} and the skip feature scaling $\sqrt{2}$ \cite{song2021scorebased}.}
	\label{fig_supp_scaling_loss}
\end{figure}

\textbf{The skip feature scaling in the Decoder.} Varying values of $t$ lead to oscillations in the loss values for DDPMs, which hinders the effective convergence of models. Some works \cite{saharia2022image, song2021scorebased, huang2022scale} suggest that adjusting the skip feature scaling in the decoder can effectively address this issue:

\vspace{-5pt}
\begin{equation}
\begin{split}
	\label{supp_f431}
        F=cat(F_{SF} \times sf, F_{BF})
\end{split}
\end{equation}
where $F_{SF}$ means the skip features from the Encoder, while $F_{BF}$ represents the backbone features from the Decoder. $sf$ indicates the scaling factor. Fig.~\ref{fig_supp_scaling_loss}(a) illustrates this detail of the skip feature scaling in the Decoder.

Tab.~\ref{supp_tab53} shows that reducing the proportion of skip features in the decoder of NN makes training more stable and enhances the generative capability of models. Meanwhile, Fig.~\ref{fig_supp_scaling_loss}(b) further supports this viewpoint, as the training loss curve becomes smoother when using this trick.

We chose the model with the skip feature scaling $\sqrt{2}$ as the baseline.

\begin{table}[h]
        \scriptsize
  \resizebox{0.48\textwidth}{!}{
	\begin{tabular}{p{1.0cm}p{0.8cm}p{0.8cm}p{0.8cm}p{0.005cm}p{1.2cm}}	
        \Xhline{1pt}
  
        \multirow{2}{*}{Methods}
        &\multicolumn{3}{c}{Performance} 
        &\quad
        &\makecell[c]{Robustness}  \\
        \cline{2-4} \cline{6-6}
        
        &\makecell[c]{mIoU}
        &\makecell[c]{mAcc}
        &\makecell[c]{allAcc}
        &\quad
        &\makecell[c]{$\tau$=0.1}\\
       \hline

        Baseline
        &\makecell[c]{{79.7}}
        &\makecell[c]{\underline{86.9}}
        &\makecell[c]{{94.1}}
        &\quad
        &\makecell[c]{{67.1}}\\

        SL \cite{huang2022scale}
        &\makecell[c]{\underline{79.8}}
        &\makecell[c]{{86.8}}
        &\makecell[c]{\underline{94.2}}
        &\quad
        &\makecell[c]{\underline{67.5}}\\

        \rowcolor{gray!20}
        $\sqrt{2}$ \cite{song2021scorebased}
        &\makecell[c]{\textbf{80.0}}
        &\makecell[c]{\textbf{87.0}}
        &\makecell[c]{\textbf{94.3}}
        &\quad
        &\makecell[c]{\textbf{67.6}}\\

        \Xhline{1pt}
        
	\end{tabular}
 }
	\caption{The result of the skip connection scaling in the Decoder on nuScenes.}
	\label{supp_tab53}

\end{table}

\textbf{The noise schedule range.} We found in our experiments that the noise schedule range is critical to the performance of CNF. This is because the noise schedule range can control the perturbation degree from NN for CN. 

Tab.~\ref{supp_tab54} shows the results of different noise schedule ranges (the baseline means the noise schedule range is [0.0001,0.02]). As mentioned in Sec.~5.6 of the main text, in sparse and perturbed scenes, we should choose a noise schedule with a smaller range.

\begin{table}[h]
        \scriptsize
  \resizebox{0.48\textwidth}{!}{
	\begin{tabular}{p{1.3cm}p{0.8cm}p{0.8cm}p{0.8cm}p{0.005cm}p{1.2cm}}	
        \Xhline{1pt}
  
        \multirow{2}{*}{Methods}
        &\multicolumn{3}{c}{Performance} 
        &\quad
        &\makecell[c]{Robustness}  \\
        \cline{2-4} \cline{6-6}
        
        &\makecell[c]{mIoU}
        &\makecell[c]{mAcc}
        &\makecell[c]{allAcc}
        &\quad
        &\makecell[c]{$\tau$=0.1}\\
       \hline

        Baseline
        &\makecell[c]{{80.0}}
        &\makecell[c]{{87.0}}
        &\makecell[c]{\underline{94.3}}
        &\quad
        &\makecell[c]{\underline{67.6}}\\

        [0.001,0.005]
        &\makecell[c]{\underline{80.3}}
        &\makecell[c]{\underline{87.3}}
        &\makecell[c]{\textbf{94.4}}
        &\quad
        &\makecell[c]{\textbf{67.7}}\\

        \rowcolor{gray!20}
        $[$0.002,0.003$]$
        &\makecell[c]{\textbf{80.5}}
        &\makecell[c]{\textbf{87.5}}
        &\makecell[c]{\textbf{94.4}}
        &\quad
        &\makecell[c]{{67.5}}\\

        \Xhline{1pt}
        
	\end{tabular}
 }
	\caption{The results of the different noise schedule ranges on nuScenes.}
	\label{supp_tab54}

\end{table}

\textbf{The loss strategy.} CNF with two branches can also be optimized from a multi-task perspective. This can further constrain unreasonable perturbations of NN, due to the convergence speed difference between NN and CN. As mentioned in Tab.~9 of the main text, we tried several loss strategies: : 1) Equal Weighting (EW, Baseline). This means that the losses of all tasks use the same weight. 2) Random Loss Weighting (RLW) \cite{lin2021reasonable}. The random weight are assigned to the losses of all tasks. 3) Uncertainty Weights (UW) \cite{kendall2018multi}. This utilizes a learnable weight to balance the losses of all tasks. 4) Geometric Loss Strategy (GLS) \cite{chennupati2019multinet++}. This mitigates the convergence differences of multiple tasks through a geometric mean weight. 

Tab.~\ref{supp_tab55} shows that GLS achieves the best results. GLS can alleviate the difference in convergence speed between NN and CN, making the noise perturbation from NN more reasonable.

\begin{table}[h]
        \scriptsize
  \resizebox{0.48\textwidth}{!}{
	\begin{tabular}{p{1.0cm}p{0.8cm}p{0.8cm}p{0.8cm}p{0.005cm}p{1.2cm}}	
        \Xhline{1pt}
  
        \multirow{2}{*}{Methods}
        &\multicolumn{3}{c}{Performance} 
        &\quad
        &\makecell[c]{Robustness}  \\
        \cline{2-4} \cline{6-6}
        
        &\makecell[c]{mIoU}
        &\makecell[c]{mAcc}
        &\makecell[c]{allAcc}
        &\quad
        &\makecell[c]{$\tau$=0.1}\\
       \hline

        Baseline
        &\makecell[c]{{80.5}}
        &\makecell[c]{{87.5}}
        &\makecell[c]{{94.4}}
        &\quad
        &\makecell[c]{{67.5}}\\

        RLW \cite{lin2021reasonable}
        &\makecell[c]{{80.4}}
        &\makecell[c]{{87.5}}
        &\makecell[c]{{94.3}}
        &\quad
        &\makecell[c]{{67.5}}\\

        UW \cite{kendall2018multi}
        &\makecell[c]{\underline{80.6}}
        &\makecell[c]{\underline{87.7}}
        &\makecell[c]{\underline{94.7}}
        &\quad
        &\makecell[c]{\underline{67.6}}\\

        \rowcolor{gray!20}
        GLS \cite{chennupati2019multinet++}
        &\makecell[c]{\textbf{80.8}}
        &\makecell[c]{\textbf{87.8}}
        &\makecell[c]{\textbf{94.8}}
        &\quad
        &\makecell[c]{\textbf{67.8}}\\

        \Xhline{1pt}
        
	\end{tabular}
 }
	\caption{The results of the different loss strategies.}
	\label{supp_tab55}

\end{table}

\vspace{-10pt}

\section{Limitations}
\label{sec6}

In the main text, CNF demonstrates excellent results across various tasks, such as segmentation and classification. However, during the design and experimentation process, we also identified some limitations of CNF.

\begin{figure}[htp]
	\centering
	\includegraphics[width=0.48\textwidth]{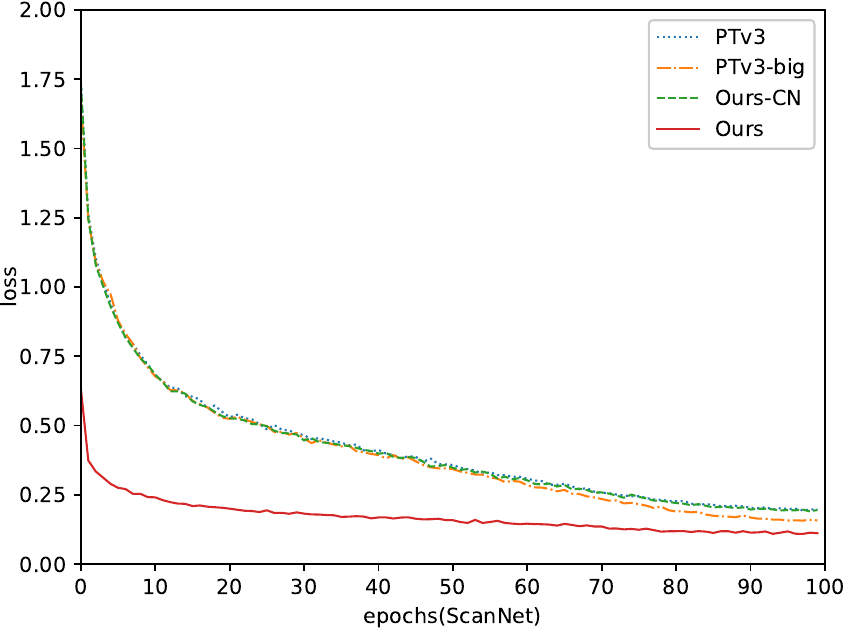}
 \vspace{-0.5cm} 
	\caption{The training loss curves of PTv3, PTv3-big, Ours-CN and Ours. PTv3-big means using the same network architecture configuration as PTv3-PPT. Our-CN represents our method with NN entirely removed, retaining only CN. The loss curves of PTv3-big and Our-CN are almost identical to that of PTv3 on ScanNet. However, PTv3-big and Our-CN, with more parameters compared to PTv3, show poorer performance, demonstrating overfitting.}
	\label{fig_supp_loss_function}
\end{figure}

\begin{itemize} 
    \item \textbf{Indirectly inheriting the robustness from DDPMs.} To avoid the extensive training and inference iterations from DDPMs, CNF uses CN as the task-dominant network. This also results in only indirectly inheriting the robustness from DDPMs. Therefore, under the same setting (e.g., both models of NCF and CNF achieve the same task performance), we believe that the robustness of CNF will be lower than that of NCF.
    \item \textbf{Limited robustness to noise.} As mentioned in Sec.~5.3 of the main text, the robustness of CNF is limited to noise from the modeled or approximate distribution, but sensitive to noise far from the modeled distribution.
    \item \textbf{Requiring more parameters.} CNF requires additional NN and FFM, resulting in requiring more parameters. Nevertheless, this also enhances the generalization of models, alleviating overfitting, as shown in Fig.~\ref{fig_supp_loss_function}. PTv3-big and Our-CN, with more parameters compared to PTv3, show lower training loss curves, but exhibit worse generalization performance on ScanNet. In comparison, our method shows the lower training loss curve and the better performance. Meanwhile, CNF can demonstrate outstanding performance in outdoor scenes, as outdoor scenes are more sparse and perturbed compared to indoor scenes. Tab.~\ref{supp_tab61} shows that PTv3, with CNF introduced and amounting to only about half the parameter count of PTv3-big, exhibits better performance.
    \item \textbf{Difficult to apply to generative tasks.} Our goal is to propose a new network framework that lowers the threshold for applying DDPMs to various 3D tasks. However, this appears to be difficult to apply to generative tasks, as the introduction of randomness in CN is indirect. This may result in a lack of diversity in the generated outcomes (see Sec.~\ref{sec3.7}). We suggest to still use NCF of DDPMs in  generation tasks. 

    
\end{itemize}

\begin{table}[h]
        \scriptsize
  \resizebox{0.48\textwidth}{!}{
	\begin{tabular}{p{1.0cm}p{0.8cm}p{0.8cm}p{0.8cm}p{0.005cm}p{1.2cm}}	
        \Xhline{1pt}
  
        \multirow{2}{*}{Methods}
        &\multicolumn{3}{c}{Performance} 
        &\quad
        &\multirow{2}{*}{Params}  \\
        \cline{2-4} 
        
        &\makecell[c]{mIoU}
        &\makecell[c]{mAcc}
        &\makecell[c]{allAcc}
        &\quad\\
        \Xhline{1pt}

	&\multicolumn{5}{c}{ScanNet} \\
        \cline{2-6}

        PTv3
        &\makecell[c]{{77.6}}
        &\makecell[c]{\underline{85.0}}
        &\makecell[c]{\underline{92.0}}
        &\quad
        &\makecell[c]{46.2M}\\

        PTv3-big
        &\makecell[c]{76.8}
        &\makecell[c]{\underline{85.0}}
        &\makecell[c]{91.8}
        &\quad
        &\makecell[c]{97.3M}\\

        Ours-CN
        &\makecell[c]{76.6}
        &\makecell[c]{84.6}
        &\makecell[c]{91.6}
        &\quad
        &\makecell[c]{88.1M}\\

        Ours-x0
        &\makecell[c]{\underline{77.7}}
        &\makecell[c]{84.7}
        &\makecell[c]{91.6}
        &\quad
        &\makecell[c]{101.4M}\\

        \rowcolor{gray!20}
        Ptv3+CNF
        &\makecell[c]{\underline{77.7}}
        &\makecell[c]{84.8}
        &\makecell[c]{91.7}
        &\quad
        &\makecell[c]{59.4M}\\

        \rowcolor{gray!20}
        Ours
        &\makecell[c]{\textbf{77.9}}
        &\makecell[c]{\textbf{85.2}}
        &\makecell[c]{\textbf{92.2}}
        &\quad
        &\makecell[c]{101.4M}\\

        \Xhline{1pt}

        &\multicolumn{5}{c}{ScanNet200} \\
        \cline{2-6}

        PTv3
        &\makecell[c]{35.3}
        &\makecell[c]{\textbf{46.0}}
        &\makecell[c]{83.3}
        &\quad
        &\makecell[c]{46.2M}\\

        PTv3-big
        &\makecell[c]{35.4}
        &\makecell[c]{45.5}
        &\makecell[c]{83.2}
        &\quad
        &\makecell[c]{97.3M}\\

        Ours-CN
        &\makecell[c]{35.3}
        &\makecell[c]{45.6}
        &\makecell[c]{83.0}
        &\quad
        &\makecell[c]{88.1M}\\

        Ours-x0
        &\makecell[c]{35.8}
        &\makecell[c]{45.4}
        &\makecell[c]{83.3}
        &\quad
        &\makecell[c]{101.4M}\\

        \rowcolor{gray!20}
        Ptv3+CNF
        &\makecell[c]{35.9}
        &\makecell[c]{45.3}
        &\makecell[c]{\underline{83.4}}
        &\quad
        &\makecell[c]{59.4M}\\

        \rowcolor{gray!20}
        Ours
        &\makecell[c]{\textbf{36.3}}
        &\makecell[c]{\underline{45.9}}
        &\makecell[c]{\textbf{83.9}}
        &\quad
        &\makecell[c]{101.4M}\\

        \Xhline{1pt}

        &\multicolumn{5}{c}{Nuscenes} \\
        \cline{2-6}

        PTv3
        &\makecell[c]{80.3}
        &\makecell[c]{87.2}
        &\makecell[c]{94.6}
        &\quad
        &\makecell[c]{46.2M}\\

        PTv3-big
        &\makecell[c]{80.4}
        &\makecell[c]{\underline{87.2}}
        &\makecell[c]{\underline{94.5}}
        &\quad
        &\makecell[c]{97.3M}\\

        Ours-CN
        &\makecell[c]{80.4}
        &\makecell[c]{87.1}
        &\makecell[c]{94.2}
        &\quad
        &\makecell[c]{88.1M}\\

        Ours-x0
        &\makecell[c]{80.7}
        &\makecell[c]{87.6}
        &\makecell[c]{94.7}
        &\quad
        &\makecell[c]{101.4M}\\

        \rowcolor{gray!20}
        PTv3+CNF
        &\makecell[c]{\underline{81.0}}
        &\makecell[c]{\textbf{87.9}}
        &\makecell[c]{\textbf{94.8}}
        &\quad
        &\makecell[c]{59.4M}\\

        \rowcolor{gray!20}
        Ours
        &\makecell[c]{\textbf{81.2}}
        &\makecell[c]{\underline{87.8}}
        &\makecell[c]{\textbf{94.8}}
        &\quad
        &\makecell[c]{101.4M}\\

        \Xhline{1pt}
        
	\end{tabular}
 }
	\caption{The results of PTv3, PTv3-big, Ours-CN and Ours on ScanNet, ScanNet200 and nuScenes. Our method achieves state-of-the-art performance on all benchmarks.}
	\label{supp_tab61}

\end{table}

\section{More Visualization Results}
\label{sec7}

\begin{figure*}[htp]
	\centering
	\includegraphics[width=\textwidth]
 {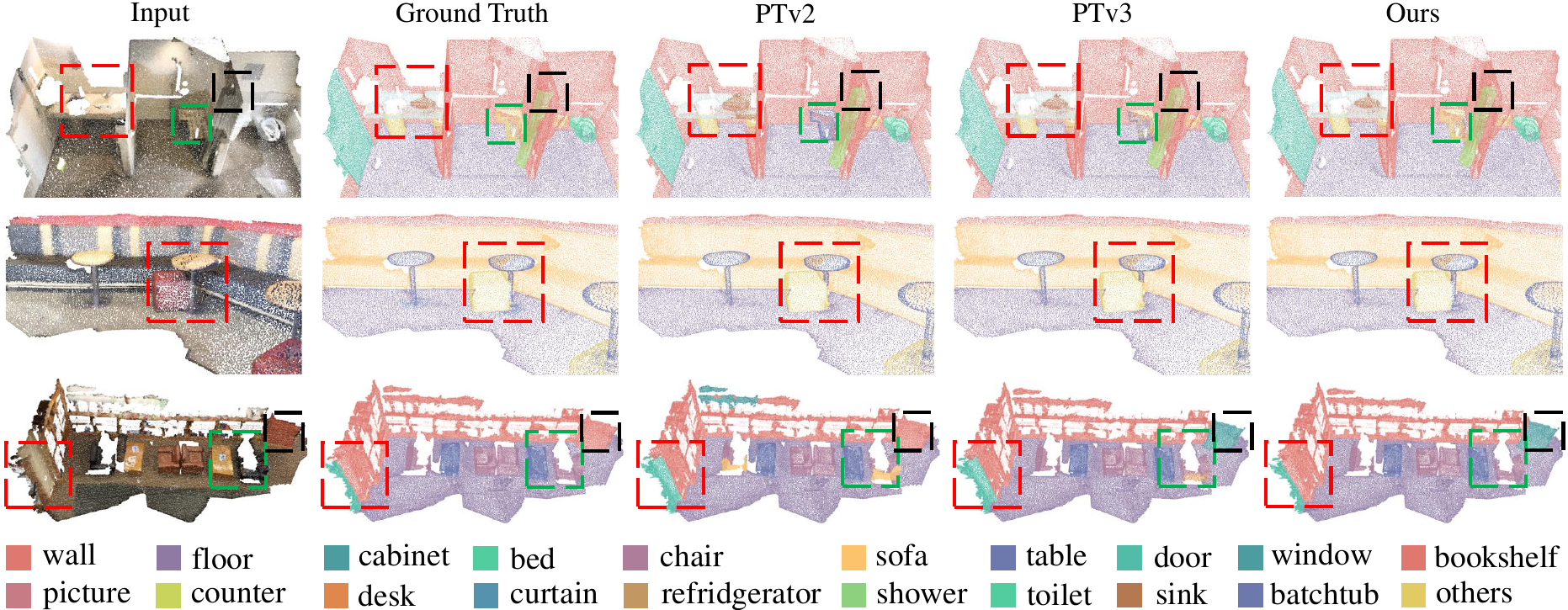}
 \vspace{-0.6cm} 
	\caption{The visualization on ScanNet. The black dashed box indicates a misannotation in the Ground Truth that we believe exists.}
	\label{fig_supp_scannet}
\end{figure*}

\begin{figure*}[htp]
	\centering
	\includegraphics[width=\textwidth]
 {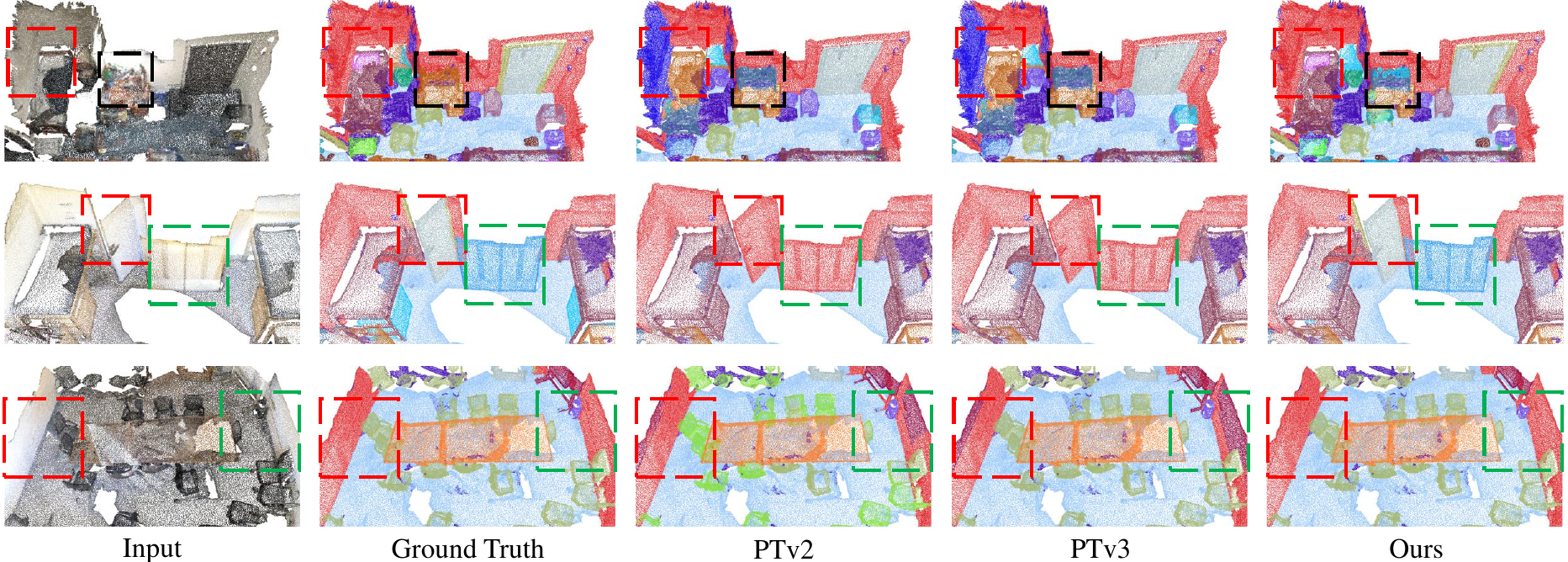}
 \vspace{-0.6cm} 
	\caption{The visualization on ScanNet200. The black dashed box indicates a misannotation in the Ground Truth that we believe exists.}
	\label{fig_supp_scannet200}
\end{figure*}

\begin{figure*}[htp]
	\centering
	\includegraphics[width=\textwidth]
 {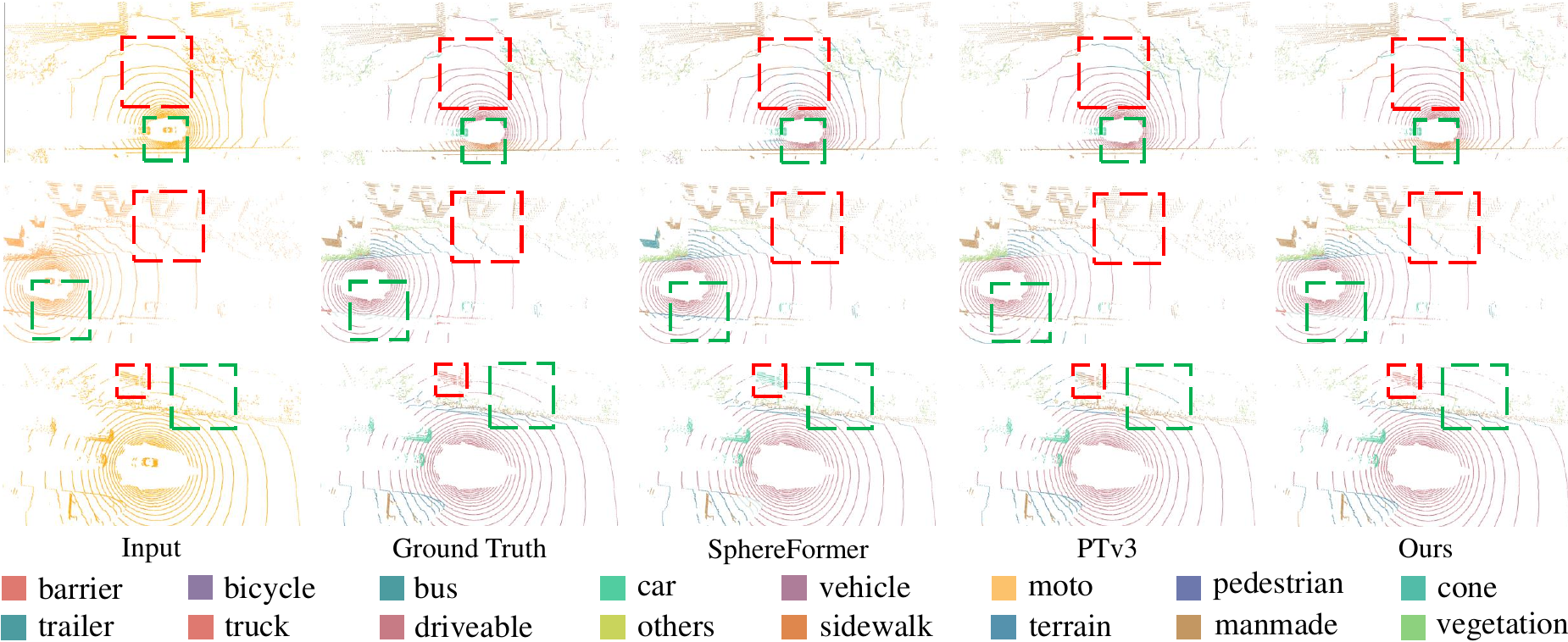}
 \vspace{-0.6cm} 
	\caption{The visualization on nuScenes.}
	\label{fig_supp_nuscenes}
\end{figure*}

We display additional visual results of semantic segmentation in Fig.~\ref{fig_supp_scannet}, Fig.~\ref{fig_supp_scannet200} and Fig.~\ref{fig_supp_nuscenes}.

\end{document}